\documentclass[sigconf]{acmart}
\usepackage{graphics}
\usepackage{graphicx}
\usepackage{xspace}
\usepackage{subcaption}
\usepackage{amsmath}
\usepackage{amsthm}
\usepackage{mathrsfs}
\usepackage{array}
\usepackage{textcomp}
\usepackage{weiwAlgorithm}
\usepackage{url}
\usepackage{diagbox}
\usepackage{color}
\usepackage{booktabs}
\usepackage{listings}
\usepackage{tcolorbox}
\usepackage{setspace}
\usepackage{multirow}
\usepackage{booktabs}    
\usepackage{tabularx}    
\usepackage{array}       
\usepackage{makecell}    
\usepackage{enumitem}

\usepackage{booktabs,tabularx,array,xcolor}
\definecolor{brandblue}{HTML}{8EA0CC}
\definecolor{brandgray}{HTML}{717271}

\usepackage[normalem]{ulem}
\useunder{\uline}{\ul}{}

\newtheorem{example}{Example}

\newtheorem{definition}{Definition}

\newcommand{\myparagraph}[1]{\vspace{1mm} \noindent \textbf{#1}.}

\newcommand{\myparagraphunderline}[1]{\vspace{0.5mm} \noindent \underline{#1.}}
\newcommand{\myparagraphquestion}[1]{\vspace{1mm} \noindent \textbf{#1?}}
\newcommand{\myparagraphunderlinenew}[1]{\vspace{0.5mm} \noindent \underline{#1,}}
\usepackage{booktabs,multirow}
\usepackage{colortbl,xcolor}
\usepackage{pgfmath}

\usepackage[normalem]{ulem} 

\newcommand{\ie}{{i.e.,}\xspace}
\newcommand{\eg}{{e.g.,}\xspace}

\newcommand{\mot}{{MemoTime}\xspace}
\newcommand{\Ind}{\mathrm{I}}

\usepackage{booktabs,multirow}
\usepackage{caption}      
\usepackage{placeins}     

\AtBeginDocument{%
  }

\copyrightyear{2026}
\acmYear{2026}
\setcopyright{cc}
\setcctype{by}
\acmConference[WWW '26]{Proceedings of the ACM Web Conference 2026}{April 13--17, 2026}{Dubai, United Arab Emirates}
\acmBooktitle{Proceedings of the ACM Web Conference 2026 (WWW '26), April 13--17, 2026, Dubai, United Arab Emirates}
\acmPrice{}
\acmDOI{10.1145/3774904.3792581}
\acmISBN{979-8-4007-2307-0/2026/04}

\begin{document}
\lstdefinelanguage{SPARQL}{
  morekeywords={BASE,PREFIX,SELECT,WHERE,FILTER,OPTIONAL,DISTINCT,GRAPH,UNION,ASK,CONSTRUCT,DESCRIBE,FROM,NAMED,ORDER,BY,ASC,DESC,LIMIT,OFFSET,BIND,VALUES},
  sensitive=false,
  morecomment=[l]{\#},
  morestring=[b]",
}

\lstset{
  language=SPARQL,
  basicstyle=\ttfamily\small,
  commentstyle=\color{green},
  keywordstyle=\color{blue},
  stringstyle=\color{red},
  breaklines=true,
  escapechar=|,
}

\title{\mot: Memory-Augmented Temporal Knowledge Graph Enhanced Large Language Model Reasoning}

\author{Xingyu Tan}
\orcid{0009-0000-7232-7051}
\affiliation{%
  \institution{UNSW}
  \institution{Data61, CSIRO}
  \city{Sydney}
  \country{Australia}}
\email{xingyu.tan@unsw.edu.au}

\author{Xiaoyang Wang}
\orcid{0000-0003-3554-3219}
\affiliation{%
  \institution{UNSW}
  \city{Sydney}
  \country{Australia}}
\email{xiaoyang.wang1@unsw.edu.au}
\authornote{Corresponding author.}

\author{Qing Liu}
\orcid{0000-0001-7895-9551}
\affiliation{%
  \institution{Data61, CSIRO}
  \city{Hobart}
  \country{Australia}}
\email{q.liu@data61.csiro.au}

\author{Xiwei Xu}
\orcid{0000-0002-2273-1862}
\affiliation{%
  \institution{Data61, CSIRO}
  \city{Sydney}
  \country{Australia}}
\email{xiwei.xu@data61.csiro.au}

\author{Xin Yuan}
\orcid{0000-0002-9167-1613}
\affiliation{%
  \institution{Data61, CSIRO}
    \institution{UNSW}
  \city{Sydney}
  \country{Australia}}
\email{xin.yuan@data61.csiro.au}

\author{Liming Zhu}
\orcid{0000-0001-5839-3765}
\affiliation{%
  \institution{Data61, CSIRO}
  \city{Sydney}
  \country{Australia}}
\email{liming.zhu@data61.csiro.au}

\author{Wenjie Zhang}
\orcid{0000-0001-6572-2600}
\affiliation{%
  \institution{UNSW}
  \city{Sydney}
  \country{Australia}}
\email{wenjie.zhang@unsw.edu.au}

\renewcommand{\shortauthors}{Xingyu Tan et al.}

\begin{abstract}
Large Language Models (LLMs) have achieved impressive reasoning abilities, but struggle with temporal understanding, especially when questions involve multiple entities, compound operators, and evolving event sequences. Temporal Knowledge Graphs (TKGs), which capture vast amounts of temporal facts in a structured format, offer a reliable source for temporal reasoning. However, existing TKG-based LLM reasoning methods still struggle with four major challenges: maintaining temporal faithfulness in multi-hop reasoning, achieving multi-entity temporal synchronization, adapting retrieval to diverse temporal operators, and reusing prior reasoning experience for stability and efficiency. To address these issues, we propose \textbf{MemoTime}, a memory-augmented temporal knowledge graph framework that enhances LLM reasoning through structured grounding, recursive reasoning, and continual experience learning. MemoTime decomposes complex temporal questions into a hierarchical Tree of Time, enabling operator-aware reasoning that enforces monotonic timestamps and co-constrains multiple entities under unified temporal bounds. A dynamic evidence retrieval layer adaptively selects operator-specific retrieval strategies, while a self-evolving experience memory stores verified reasoning traces, toolkit decisions, and sub-question embeddings for cross-type reuse. Comprehensive experiments on multiple temporal QA benchmarks show that MemoTime achieves overall state-of-the-art results, outperforming the strong baseline by up to 24.0\%. Furthermore, MemoTime enables smaller models (e.g., Qwen3-4B) to achieve reasoning performance comparable to that of GPT-4-Turbo.

\end{abstract}



\maketitle

\vspace{-4mm}
\section{Introduction}
\label{sec:intro}
\begin{figure}[t]
    \centering
    \includegraphics[width=0.98\linewidth]{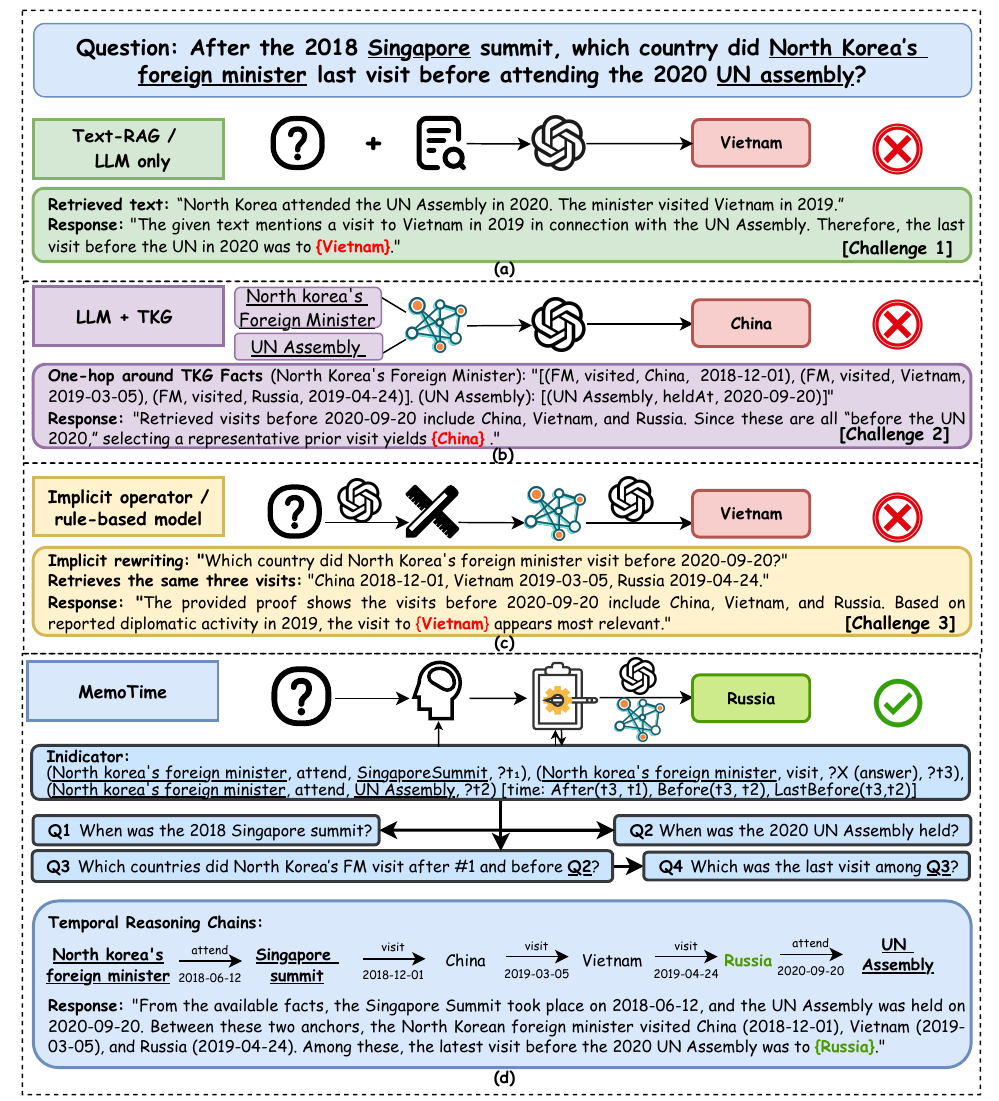}
    \vspace{-5pt}
    \caption{Representative workflow of four LLM reasoning paradigms.}
    \label{fig:intro_demo}
    \vspace{-5pt}
    
\end{figure}

Large Language Models (LLMs) have demonstrated strong performance across a wide range of tasks by scaling to billions of parameters and pre-training on massive and diverse text corpora~\cite{brown2020language}. 
However, due to the prohibitive cost of retraining, these models are inherently static.
This limitation results in factual gaps, temporal inconsistencies, and reasoning hallucinations when LLMs are required to process time-sensitive or evolving information~\cite{petroni2020kilt, talmor2018commonsenseqa,khot2022decomposed}.

Retrieval-Augmented Generation (RAG) has emerged as an effective paradigm to mitigate these limitations by allowing LLMs to access external information sources during inference~\cite{ huang2025survey, huang2025embedding,zheng2025lifting}. 
Typical RAG systems embed both questions and documents into a shared vector space and retrieve semantically similar passages~\cite{tog2.0ma2024think, tan2025hydra}. 
Although effective in many cases, such retrieval often prioritizes semantic similarity while neglecting structural and temporal dependencies across entities and events \cite{jia2024faithful}. 
For instance, two temporally distinct statements such as “Barack Obama is the President of the United States (2009–2017)” and “Joe Biden is the President of the United States (2021–)” may appear semantically related, yet they cannot simultaneously hold true. 
Plain-text retrieval often fails to distinguish between such temporally exclusive facts, leading to incorrect or contradictory reasoning chains.

Temporal reasoning further amplifies these challenges. 
Real-world entities and relationships evolve continuously, i.e., people change roles, organizations merge, and events unfold in sequence over time \cite{liang2023learn, ARI}. 
Temporal-aware questions therefore often involve implicit time constraints, multiple dependent conditions, and mixed temporal granularities such as “Who chaired the committee before the 2010 reform?” or “Which country hosted the first summit after 2015?”. 
These questions require reasoning that integrates both temporal and semantic alignment, which remains difficult for text-based RAG methods that lack structured temporal awareness.

To address this issue, integrating Temporal Knowledge Graphs (TKGs) with LLM reasoning provides a promising direction \cite{jia2024faithful,T4, ARI, khot2022decomposed, qianyihu-etal-2025-time}. 
TKGs encode factual knowledge as quadruples (subject, relation, object, timestamp), offering explicit temporal grounding and relational structure. 
By leveraging TKGs, models can reason over evolving entities while maintaining factual and temporal consistency. 
Temporal Knowledge Graph Question Answering (TKGQA) serves as a representative evaluation for this task, requiring systems to answer natural language questions by retrieving temporally relevant facts from TKGs.

\myparagraph{Challenges in existing methods}
Most existing TKG-based reasoning frameworks follow a ``plan-retrieve-answer'' pipeline. 
In this paradigm, LLMs decompose complex temporal questions into a series of sub-tasks, retrieve related facts from a TKG, and generate an answer based on the retrieved context. 
While this improves interpretability and modularity, several challenges face.

\myparagraphunderline{Challenge 1: Temporal faithfulness in multi-hop reasoning}  
Most temporal QA pipelines\cite{jia2024faithful,T4, ARI}, as shown in Figure \ref{fig:intro_demo}(a), prioritize semantic similarity over chronological accuracy.  
Expanding from one-hop neighbors and ranking by semantics often retrieves paths that appear contextually relevant but violate time constraints.  
For example, answering “Which country did X last visit before 2020?” may simply return (X, visit, Y, 2019) because of lexical overlap, despite contradicting the before-last relation.  
In multi-hop queries such as “Who did X hire after Y resigned during Q3 2017?”, locally valid hops can combine into a globally inconsistent reasoning chain.

\myparagraphunderline{Challenge 2: Multi-entity temporal synchronization}  
When a question contains multiple entities, most systems \cite{T4,  tog1.0sun2023think,plan-on-graph} explore each entity independently and attempt to merge partial evidence later.  
This disjoint process often produces candidates that never align within a single, time-consistent reasoning path.  
For instance, as shown in Figure \ref{fig:intro_demo}(b), independent exploration can yield valid facts for each entity, but fails to synchronize their temporal windows.  

\myparagraphunderline{Challenge 3: Operator diversity and adaptive retrieval}  
Temporal questions encompass diverse operators, each requiring a distinct reasoning policy.  
Previous methods \cite{saxena2021question,T4} focus on 
single-round rewriting and can expose explicit timestamps but struggle with combined constraints, as shown in Figure \ref{fig:intro_demo}(c).  
It prevents the model from adapting to such heterogeneity, resulting in either under-coverage (missing valid evidence) or over-retrieval (noisy results).  

\myparagraphunderline{Challenge 4: Lack of reasoning experience management}  
Most existing pipelines \cite{qianyihu-etal-2025-time, T4, tog2.0ma2024think, pogtan2025paths} remain memoryless, discarding successful reasoning traces after each run.  
They often rely on manually crafted or static exemplars, which are costly to construct and insufficiently generalizable across diverse temporal question types, leading models repeatedly re-solve similar sub-questions from scratch, failing to transfer prior knowledge across operators or tasks.  

\myparagraph{Contribution} 
In this paper, we introduce \textbf{\mot}, a \text{Memory-Augmented Temporal Knowledge Graph} framework designed to enhance LLM temporal reasoning, as shown in Figure \ref{fig:intro_demo}(d).  
Unlike existing methods that rely on static retrieval or pre-defined templates, \mot integrates structured temporal grounding, hierarchical reasoning, dynamic toolkit invocation, and continual memory updating into a unified framework.  
It enables LLMs to reason faithfully over time-aware facts, adapt retrieval strategies to temporal operators, and progressively improve performance through experience reuse.  


\myparagraphunderlinenew{To address multi-hop temporal reasoning problem}  
\mot introduces a hierarchical reasoning framework that decomposes complex temporal questions under a unified global plan. 
All sub-indicators are evolved from the root indicator of the main question, ensuring that each branch inherits consistent temporal constraints and preserves monotonic timestamp progression.  
Guided by this global supervision, \mot performs controlled branch expansion, retrieving multi-hop reasoning paths remaining faithful to both semantic relevance and chronological order.  

\myparagraphunderlinenew{To handle multi-entity temporal synchronization}  
\mot ensures that the final reasoning path jointly incorporates all topic entities under a unified temporal framework and synchronized timeline.  
Instead of exploring entities independently, the system retrieves co-constrained evidence paths that preserve shared temporal bounds and consistent granularity, ensuring global coherence and containing the factual basis for the correct answer.

\myparagraphunderlinenew{To adapt to diverse temporal operators}  
\mot introduces a library of temporal reasoning toolkits that support heterogeneous operators.
Instead of fixed templates, \mot adaptively selects the most suitable toolkit through experience-guided prompts.
This adaptive retrieval policy ensures that each temporal operator triggers an appropriate reasoning strategy, thereby improving precision and reducing over-retrieval noise.

\myparagraphunderlinenew{To manage and reuse reasoning experience}  
\mot maintains a continuously evolving experience memory that records successful records.  
Each experience entry is stored alongside embeddings of both the question and its temporal indicator, enabling efficient similarity-based retrieval under operator and type constraints.  
During inference, \mot retrieves relevant exemplars to guide new reasoning, while after execution, verified trajectories are written back. 
The advantage of \mot can be abbreviated as follows:
\begin{itemize}[leftmargin=*]

\item \textbf{Memory-augmented temporal reasoning.} 
MemoTime introduces a unified framework that integrates dynamic memory retrieval and update into temporal-aware question decomposition,  
allowing the model to recall, reuse, and refine reasoning trajectories for long-term temporal understanding.
\item \textbf{Hierarchical and interpretable control.}  
A hierarchical controller Tree of Time executes, verifies, and refines sub-questions, ensuring faithful and interpretable temporal reasoning.

\item \textbf{Hybrid retrieval and pruning.}  
An operator-aware retrieval layer combines symbolic graph expansion with embedding search, applying temporal-first pruning and semantic re-ranking for 
both chronological validity and precise evidence grounding.
\item \textbf{Self-evolving experience memory.}  
Verified reasoning responses are continuously organized in an adaptive experience pool, forming a closed feedback loop for continual improvement.

\item\textbf{Efficiency and adaptability}:  
a) \mot is a plug-and-play framework that can be seamlessly applied to various LLMs and TKGs.
b) \mot is auto-refresh. New information is incorporated instantly via TKG retrieval instead of costly LLM fine-tuning.
c) \mot achieves state-of-the-art results on all the tested datasets, surpasses the strong baseline by up to 24.0\%, and enables smaller models (e.g., Qwen3-4B) to achieve reasoning performance comparable to GPT-4-Turbo.
\end{itemize}

\section{Related Work}
\label{sec:related_work}
\myparagraph{LLM-based knowledge graphs reasoning}
Graphs are a natural representation for modeling relational structure among entities \cite{li2024adarisk,li2025efficient,wang2025time, wang2025effective, wang2024efficient,wang2024simpler,wang2023towards, tan2023higher, zhai2025sgpt, zhai2025graph}.
Knowledge graphs (KGs) provide structured, verifiable knowledge that complements the implicit world knowledge in LLMs~\cite{pan2024unifying, guan2024mitigating, wu2023holistic,yanpingwu, xie2025hl}. 
Early works embedded KG facts into neural networks during pre-training or fine-tuning~\cite{zhang2021poolingformer,rogluo2023reasoning,sima2025deep,sima2026beyond, zheng2024understanding}, 
but such approaches hinder efficient updates and reduce interpretability. 
Recent studies explore LLM–KG integration by prompting LLMs to iteratively traverse graphs, as seen in  ~\cite{tog1.0sun2023think,jiang2023structgpt, tan2026privgemo, zai2025proh}.
These systems guide the LLM to expand reasoning paths from a seed entity and refine answers through repeated retrieval–generation cycles. 
However, starting from a single vertex overlooks multi-entity connections and temporal dependencies, 
often yielding semantically plausible yet chronologically inconsistent paths.  
\cite{pogtan2025paths} alleviates this by modeling multi-hop reasoning paths, 
but relies on static KGs and lacks the dynamic temporal alignment.

\myparagraph{Temporal knowledge graphs question answering}
Temporal Knowledge graphs extend static graphs with timestamped facts, enabling reasoning over evolving entities and relations\cite{zhang2024tatkc, zhang2024towards, yang2023hgmatch}.  
Earlier approaches fall into two main categories: semantic parsing-based and embedding-based methods.  
Parsing-based methods translate natural-language questions into logical forms executable on TKGs \cite{TEQUILA,SYGMA, Prog-TQA}.  
These achieve precise execution but fail on long or ambiguous queries due to brittle symbolic parsing.  
Embedding-based methods ~\cite{saxena2021question, mavromatis2022tempoqr, mavromatis2022tempoqr} learn vectorized temporal reasoning by aligning question embeddings with fact embeddings, but they are limited to short reasoning chains and simple time expressions.  
Recent LLM-based systems~\cite{jia2024faithful,gao2024two,T4} improve interpretability by generating reasoning steps in natural language.

\myparagraph{Hierarchical and memory-augmented reasoning}  
Recent advances in LLM-based QA frameworks have emphasized question decomposition as a way to improve reasoning depth and interpretability.  
Multi-hop reasoning systems such as~\cite{yao2022react} decompose complex queries into simpler sub-questions that are independently solved and aggregated.  
This decomposition enables finer-grained retrieval and targeted reasoning across heterogeneous evidence sources~\cite{feldmanMultiHopParagraphRetrieval2019}.  
However, these methods decompose linearly, leading to accumulated errors.  
Moreover, most existing decomposition modules are either manually designed~\cite{minMultihopReadingComprehension2019} or fine-tuned~\cite{wuGenDecRobustGenerative2024} for specific datasets, limiting their adaptability across temporal and structural question types.  
In addition to the decomposition, recent studies explore long-term memory systems for LLM agents~\cite{aios}.  
Approaches such as ~\cite{memorybank} enable retrieval of prior interaction histories, while SCM~\cite{wang2023enhancing} maintains selective access through controller mechanisms.  
But these frameworks are typically task-agnostic and lack structured representations of reasoning processes.

\begin{figure*}[h]
    \centering
    \includegraphics[width=0.85\linewidth]{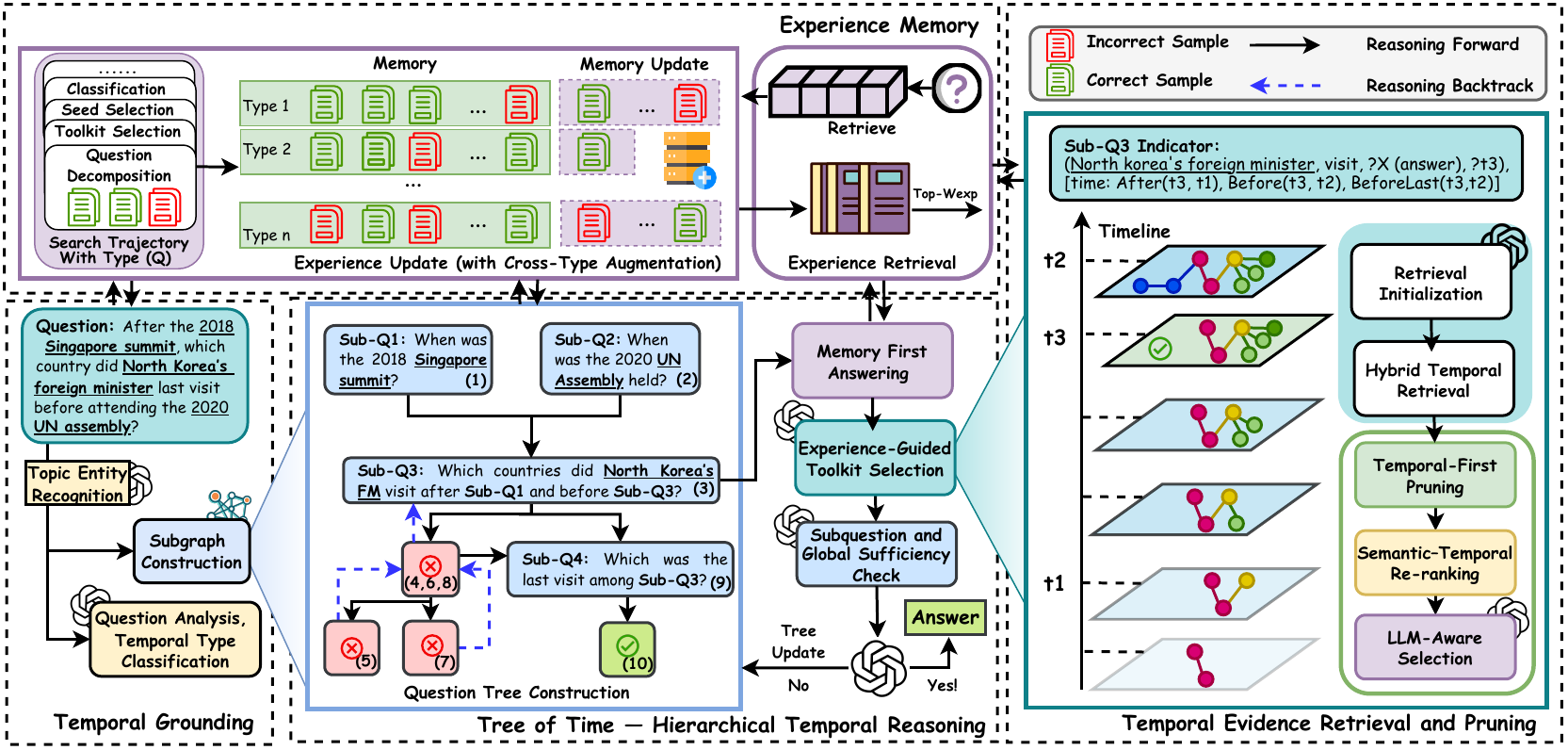}
    \caption{ 
    Overview of the MemoTime framework.
Temporal Grounding: Topic entities and temporal operators are extracted from the input question to construct a question-specific subgraph.
Tree of Time 
(Hierarchical Reasoning): 
Recursively decomposes the question into sub-questions guided by temporal dependencies, adaptively reusing experience or invoking toolkits for new evidence.
Temporal Evidence Retrieval and Pruning:  Performs operator-aware retrieval under monotonic time constraints, followed by semantic-temporal re-ranking and LLM-based sufficiency verification.
Experience Memory: Verified reasoning traces are stored, updated, and retrieved for cross-type reuse, enabling continual self-improvement across reasoning cycles.
}
    \label{fig:overall}
    \vspace{-1em}
\end{figure*}

\vspace{-2mm}
\section{Preliminaries}
\label{sec:preliminaries}
Consider a Temporal Knowledge Graph (TKG) $\mathcal{G} = (\mathcal{E}, \mathcal{R}, \mathcal{T})$, where $\mathcal{E}$, $\mathcal{R}$, and $\mathcal{T}$ represent the set of entities, relations, and timestamps, respectively. 
$\mathcal{G} = (\mathcal{E}, \mathcal{R}, \mathcal{T})$ contains abundant temporal factual knowledge in the form of quadruples, i.e., $\mathcal{G} = \{(e_h, r, e_t, t) \mid e_h,e_t \in \mathcal{E}, r \in \mathcal{R}, t \in \mathcal{T} \}$.
%
Temporal constraint defines a condition related to a specific time point or interval that must be satisfied by both the answer and its supporting evidence. Following Allen's interval algebra~\cite{allen1984towards}, this includes the 13 classical temporal relations, as well as extended operators such as temporal set relations, duration comparisons, and sorting mechanisms (e.g., first, last, nth)~\cite{TimelineKGQA}. Temporal constraints thus enable fine-grained alignment between natural language queries and temporally consistent evidence paths.

\begin{definition}[Temporal Path] 
Given a TKG $\mathcal{G}$, a temporal  path is a connected sequence of temporal facts, represented as:  
$
path_{\mathcal{G}}(e_1, e_{l+1}) = \{(e_1, r_1, e_2, t_1), (e_2, r_2, e_3, t_2), \ldots, (e_l, r_l, e_{l+1}, t_l)\},$ where $l$ denotes the length of the path, i.e., ${\rm{length}}({\rm{path}}_{\mathcal{G}}(e_1, e_{l+1})) = l$. The path must satisfy two constraints:  
(1) connectivity: the tail entity of each fact is identical to the head entity of the next;  
(2) temporal monotonicity: timestamps are non-decreasing, \ie $t_1 \leq t_2 \leq \cdots \leq t_l$.  

\end{definition}

\begin{example}[Temporal Path]
Consider a temporal path between ``Merkel'' and ``EU'' with a length of 3 :
$path_{\mathcal{G}}(\text{Merkel}, \text{EU}) = \{
(\text{Merkel},~\textit{visit},~\text{Paris},~2012),~
(\text{Paris},~\textit{host},~\text{Conference},~2013),$ $
(\text{Conference},~\textit{attended\_by},~\text{EU},~2014)
\}$,
and can be visualized as:
\[
\text{Merkel} \xrightarrow[\text{2012}]{\textit{visit}} 
\text{Paris} \xrightarrow[\text{2013}]{\textit{host}} 
\text{Conference} \xrightarrow[\text{2014}]{\textit{attended\_by}} 
\text{EU}.
\]
\end{example}

\begin{definition}[Temporal Reasoning Path]  
Given a TKG $\mathcal{G}$, and an entity list $[e_1,e_2,\ldots,e_n]$, a temporal reasoning path is a sequence of temporal segments
$
TRP_{\mathcal{G}}([e_1,\ldots,e_n]) = \{P_1,P_2,\ldots,P_{n}\},
$
where each $P_i = path_{\mathcal{G}}(e_i,n_i)$ for some $n_i \in \mathcal{E}$,  
and the global monotonicity condition holds:
$
t^{end}(P_i) \leq t^{start}(P_{i+1}), \quad \forall~1 \leq i < n.
$
\end{definition}

\begin{example}[Temporal Reasoning Path]
Consider a temporal reasoning path $TRP_{\mathcal{G}}([{Obama}, {Beijing}, {Paris}])$, by two temporally aligned segments:
$P_1$= \{(Obama, meet, UN, 2009)\}, $P_2$= \{(Beijing, linked\_via, EU, 2011), (EU, event\_in, Paris, 2012)\}, visualized as:
\[
\text{Obama} \xrightarrow[\text{2009}]{\textit{meet}} 
\text{UN} 
\;\;\leadsto\;\;
\text{Beijing} \xrightarrow[\text{2011}]{\textit{linked\_via}} 
\text{EU} \xrightarrow[\text{2012}]{\textit{event\_in}} 
\text{Paris}.
\]
\end{example}

\noindent
Temporal Knowledge Graph Question Answering (TKGQA)
is a fundamental reasoning task based on TKGs. Given a natural language question $Q$ and a TKG $\mathcal{G}$, the objective is to devise a function $f$ that identify answer entities or timestamps $a \in Answer(Q)$ utilizing knowledge in $\mathcal{G}$, \ie $a = f(q, \mathcal{G})$.
Consistent with previous research \cite{sun2019pullnet, rogluo2023reasoning, tog1.0sun2023think, tog2.0ma2024think},
we assume topic entities $Topics$ are mentioned in $Q$ and answer entities or timestamps  $Answer(q)$ in ground truth are linked to $\mathcal{G}$, i.e., $Topics \subseteq \mathcal{E} \text{ and }  Answer(Q) \subseteq \{\mathcal{E} \cup \mathcal{T}\}$.

\section{Method}
\label{sec:method}

\noindent\textbf{Overview.}
\mot\ implements ``TKG-grounded LLM reasoning'' by grounding each question in temporal facts, recursively decomposing it into executable sub-queries, retrieving and pruning temporally valid evidence, and continually refining its experience memory. 
Unlike prior TKGQA or RAG systems that depend on static retrievers or fixed prompts, \mot\ integrates temporal alignment, hierarchical control, and self-evolving memory into a unified reasoning framework. 
The overall architecture of \mot is detailed in Figure~\ref{fig:overall}, consisting of four key components:

\begin{itemize}[leftmargin=*]
  \item \textbf{Temporal Grounding.} 
  The model begins by linking topic entities and constructing a $D_{\max}$-hop temporal subgraph $\mathcal{G}_Q$ from the TKG. 
  It then classifies the temporal type of the question using exemplars retrieved from memory, producing structured temporal constraints for downstream reasoning (Section~\ref{sec:method:grounding}).

  \item \textbf{Tree of Time Reasoning.} 
  Given the grounded question and its temporal type, \mot\ constructs a hierarchical decomposition tree and executes sub-questions in a top-down manner. 
  For each node, it dynamically decides whether to recall prior experiences, invoke temporal toolkits, or refine unresolved branches, ensuring temporal consistency and interpretability (Section~\ref{sec:recursion}).

  \item \textbf{Temporal Evidence Retrieval and Pruning.} 
  Guided by toolkit configurations, \mot performs hybrid retrieval, combining time-monotone, graph exploration, and embedding-based search. 
  Retrieved candidates are filtered by temporal constraints, re-ranked by semantic and temporal proximity, and finally verified through an LLM-aware selection step (Section~\ref{sec:retrieval}).

  \item \textbf{Experience Memory.} 
  Verified reasoning traces, toolkit selections, and sub-question embeddings are stored in a dynamic memory pool. 
  At inference, similar experiences are retrieved by type and relevance; after reasoning, new traces are written back to continuously refine retrieval and decision-making across future questions (Section~\ref{sec:memory}).
\end{itemize}
\vspace{-4mm}
\subsection{Temporal Grounding}
\label{sec:method:grounding}

The temporal grounding stage transforms a natural-language temporal question into a structured reasoning representation, establishing a factual and temporal foundation for subsequent retrieval and inference.  
It consists of two core phases: knowledge fact grounding and question analysis.  The pseudo-code of the temporal grounding is detailed in Algorithm \ref{algorithm:analysis} of Appendix \ref{appendix:alg}.

\myparagraph{Knowledge fact grounding}
Given an input question $Q$, \mot first identifies the relevant temporal subgraph $\mathcal{G}_Q$ within the underlying TKG.  
This subgraph captures entities, relations, and time-stamped triples that are semantically or temporally associated with $Q$ within a bounded $D_{\max}$-hop neighborhood, 
serving as the factual evidence base for downstream reasoning.

\myparagraphunderline{Topic entity recognition}
To locate question-relevant entities, \mot employs LLMs to extract potential entity mentions and associated timestamps from $Q$.  
Following extraction, a Dense Retrieval Model (DRM) aligns these mentions with KG entities via embedding-based similarity matching.  
Specifically, both question keywords and KG entities are encoded into dense embeddings, and an entity index is constructed using FAISS \cite{douze2024faiss}.  
Cosine similarity is then computed between the two embedding spaces, and the top-ranked entities are selected to form the topic entity set $Topics$.  
These topic entities act as the initial anchors for constructing a localized subgraph.

\myparagraphunderline{Subgraph construction}
Once the topic entities are identified, \mot constructs a $D_{\max}$-hop subgraph $\mathcal{G}_Q$ that aggregates all triples connected to each topic entity within the defined temporal window.  
This subgraph provides a compact yet semantically rich neighborhood for temporal reasoning.  
To reduce redundancy and computational overhead, we apply graph reduction and relation clustering techniques following~\cite{pogtan2025paths}, ensuring $\mathcal{G}_Q$ remains concise while preserving the most relevant temporal relations.  

\myparagraph{Question analysis}
After constructing $\mathcal{G}_Q$, the next step is to analyze the temporal intent of the question.  
Unlike conventional TKGQA approaches that treat all questions uniformly, or use the label from the dataset, leveraging the flexibility reduces. Temporal questions exhibit diverse syntactic forms and intricate time constraints.  
To capture these nuances, \mot first performs {temporal type classification}, identifying the temporal operator that governs the question’s reasoning logic.

\myparagraphunderline{Temporal type classification}
Temporal type classification determines the intrinsic temporal relations implied by a question.  
Following Allen’s interval algebra~\cite{allen1984towards}, we consider thirteen fundamental temporal relations,  
as well as extended operators such as set relations, duration comparisons, and ordering mechanisms~\cite{TimelineKGQA}, as introduced in Section \ref{sec:preliminaries}.  
Existing methods often rely on static prompts or manually designed exemplars, limiting their adaptability to complex or unseen temporal structures.  
In contrast, \mot leverages a continuously updated experience pool $\mathcal{E}_{pool}$ that stores successful reasoning trajectories, toolkit configurations, and classification results.  
This memory-based design enables dynamic retrieval of contextually aligned exemplars, improving temporal reasoning consistency across question types.  
Formally, for a given question $Q$, type-specific exemplars are retrieved as:
$
\mathcal{E}_{\text{type}} = \textsf{GetTypeExp}(\mathcal{E}_{pool}, Q, W_{\text{exp}}),
$
where $W_{\text{exp}}$ denotes the exemplar retrieval limit.  
These exemplars are then incorporated into the LLM classification prompt:
$
\textsf{Type}(Q) = \textsf{Prompt}_{\text{TypeSelect}}(Q, \mathcal{E}_{\text{type}}),
$
allowing the model to identify the most appropriate temporal operator for $Q$.  
This exemplar-guided strategy enables the \mot to adapt to new question patterns, enhances type prediction accuracy, and lays the foundation for the subsequent decomposition stage.

\vspace{-4mm}
\subsection{Tree of Time — Hierarchical Temporal Reasoning}
\label{sec:recursion}

After temporal grounding, \mot\ enters the Tree of Time (ToT) stage,
which transforms grounded temporal structures into a hierarchical reasoning tree.
Serving as the framework’s central controller, it coordinates decomposition, toolkit execution, sufficiency testing, and answer synthesis.
Guided by the global plan, the model adaptively alternates between recalling experience, invoking operator-specific toolkits, and refining sub-questions to maintain temporal and semantic consistency.
Due to the space limitation, 
the pseudo-code of the ToT is detailed in Algorithm \ref{algorithm:controller} of Appendix \ref{appendix:alg}.

\myparagraph{Question tree construction}
Given a temporal question $Q$ and its classified temporal type $\textsf{Type}(Q)$, 
\mot\ begins by constructing a hierarchical decomposition tree $\mathcal{T}_Q$.  
Each node represents a sub-question with a specific temporal relation or constraint.  
To enhance robustness, decomposition exemplars are retrieved from the experience pool $\mathcal{E}_{pool}$ and integrated into the LLM prompt:
\[
\mathcal{E}_{\text{decomp}} = \textsf{GetDecompExp}(\mathcal{E}_{pool}, Q, W_{\text{exp}}, \textsf{Type}(Q)),
\]
\[
\mathcal{T}_Q = \textsf{Prompt}_{\text{Decompose}}(Q, \mathcal{E}_{\text{decomp}}, \textsf{Type}(Q)).
\]
This experience-guided process enables the system to build a type-aligned reasoning tree that reflects temporal dependencies among sub-questions. 
For any parent–child pair $(q_i, q_j)$, the temporal order satisfies $t(q_i) \le t(q_j)$, ensuring temporal monotonicity.

\myparagraphunderline{Indicator extraction}
From $\mathcal{T}_Q$, the system extracts a set of sub-question indicators $\{\Ind_i\}$, i.e., $\Ind_i = \langle x?, R, y?, \mathcal{C}_{time}\rangle$,
each encoding the entities, relations, and temporal constraints required for reasoning.  
Based on this, a predicted depth $D_{\text{pred}}(q_i)$ is calculated, defined as the maximum distance between the predicted answer and each topic entity.
For instance, in Figure~\ref{sec:intro}(d), the predicted depth of the overall indicator is 2.


\myparagraph{Hierarchical execution control}
The reasoning tree is traversed in a top-down manner.  
At each node, \mot\ first checks whether a similar reasoning trace exists in the experience pool for direct reuse. 
If no match or evidence insufficient, the system proceeds to dynamic toolkit selection to discover new evidence.

\myparagraphunderline{Experience-guided toolkit selection}
When memory lookup fails, \mot\ retrieves prior examples of successful toolkit usage to enhance selection accuracy:
\[
\mathcal{E}_{\text{tool}} = \textsf{GetToolkitExp}(\mathcal{E}_{pool}, q_i, \Ind_i, \textsf{Type}(Q), W_{\text{exp}}),
\]
and constructs a context-enriched prompt:
$
\mathcal{T}_i = 
\textsf{Prompt}_{\text{ToolkitSelect}}$ $
(\mathcal{T}_\theta,\, \Ind_i,\, q_i,\, \mathcal{E}_{\text{tool}}).
$
The LLM may recommend multiple toolkits simultaneously, 
each representing a distinct retrieval or reasoning strategy (\eg event ordering, interval comparison, or timeline construction).  
All selected tools are executed in parallel, and their candidate results are passed to the sufficiency evaluation stage.

\myparagraph{Question answering}
After each sub-question $q_i$ obtains its candidate temporal results, 
\mot\ evaluates whether these results sufficiently answer the local query and contribute to the overall question. 
This stage integrates evidence evaluation, global reasoning synthesis, and hierarchical update for unresolved nodes.

\myparagraphunderline{Sub-question evaluation}
For each node in the reasoning tree, the LLM assesses retrieved temporal paths for both semantic relevance and temporal validity. 
A sub-question is considered sufficient when one or more candidate paths provide consistent evidence aligned with its indicator. 
If multiple valid results exist, a lightweight debate–vote strategy aggregates them into a single coherent answer. 
The solved node and its reasoning trace are recorded in the experience pool $\mathcal{E}_{pool}$ for future reuse.

\myparagraphunderline{Global sufficiency and answer synthesis}
After all sub-questions are processed, \mot\ checks the temporal and logical coherence of the entire reasoning tree. 
When global sufficiency is achieved, verified sub-paths are merged into a unified temporal reasoning chain that summarizes the full evidence flow. 
This concise reasoning trace is then used to prompt the final LLM for answer generation, ensuring interpretability and temporal consistency.

\myparagraphunderline{Tree update and termination}
If a sub-question fails sufficiency verification, \mot\ adaptively decides whether to retry retrieval, refine the query, or perform further decomposition based on its current reasoning. 
Newly generated nodes are appended to the tree and processed recursively until all branches are solved or the maximum depth $D_{\max}$ is reached. 
Finally, verified reasoning paths are consolidated into the final evidence chain for answer generation.
\vspace{-2mm}
\subsection{Temporal Evidence Retrieval and Pruning}
\label{sec:retrieval}
As discussed in Section~\ref{sec:intro}, identifying reasoning paths that connect all topic entities is crucial for deriving accurate answers.  These paths function as interpretable chains of thought.
In this section, we operationalize the selected toolkits to perform hybrid retrieval and pruning, 
balancing {efficiency} and {accuracy} while ensuring both high temporal alignment and broad semantic coverage.
The pseudo-code of this section is detailed in Algorithm \ref{algorithm:execution} (Appendix\ref{appendix:alg}).


\myparagraphunderline{Retrieval initialization}
Before path exploration begins, \mot\ performs seed selection to identify relevant entities that serve as starting points for reasoning.  
If prior successful seeds exist in $\mathcal{E}_{pool}$, they are reused directly; otherwise, new seeds are chosen via the experience-augmented prompt:
$
S_i = \textsf{Prompt}_{\text{SeedSelect}}(Topics, \Ind_i, q_i, \mathcal{E}_{\text{seed}}),
$
where $\mathcal{E}_{\text{seed}}$ denotes the retrieved relevant examples  from memory.  

\myparagraph{Toolkit-driven temporal path retrieval}
Each toolkit $T_\theta \in \mathcal{T}_i$ represents a specialized retrieval operator for a temporal reasoning pattern 
(e.g., one-hop expansion, interval comparison, event ordering, or range detection).  
Guided by the indicator $\Ind_i$, 
each toolkit constrains both the semantic relation and temporal boundary of the retrieval.  
Multiple toolkits are executed in parallel to form a diverse set of candidate paths $\mathcal{P}_i$. 

\myparagraphunderline{Hybrid temporal retrieval}
For graph-based retrieval, the system performs time-monotonic path expansion. 
Instead of starting from the maximum depth $D_{\max}$, \mot  begins exploration at the predicted depth $D_{\text{pred}}(\Ind_i)$. Given the subgraph $\mathcal{G}_Q$, the ordered seed set $S_i$, the indicator $\Ind_{\text{i}}$, and the depth $D = \min(D_{\text{pred}}(\Ind_i), D_{\max})$, we identify candidate temporal paths that include all seeds in order. To avoid exhaustive search, we apply a tree-structured  bidirectional breadth-first search (BiBFS) from each entity to extract all potential paths, defined as:
$
C = \{p \mid |S_i| \cdot (D{-}1) < \operatorname{length}(p) \leq |S_i| \cdot D\}.$
In parallel, an embedding retriever retrieves semantically relevant facts from document-indexed KG segments, 
ensuring that implicit or cross-hop relations are also captured.  
The two result streams are concatenated and unified into a hybrid candidate path pool. 

\noindent
\textbf{Temporal and semantic pruning.}
Candidate paths are then filtered through a multi-stage pruning pipeline by temporal consistency, semantics, and LLM selection.
Together, these steps drastically reduce noise while retaining only candidates that satisfy both chronological and semantic alignment.

\myparagraphunderline{Temporal-first pruning}
Traditional relevance-first pruning may retain semantically similar but temporally inconsistent paths.  
To mitigate this, \mot\ adopts a \emph{temporal-first} pruning policy.  
Given the candidate pool $\mathcal{C} = \{p_i\}_{i=1}^{N}$ returned by the hybrid retriever,  
\mot\ first applies a strict temporal filter that removes paths violating time constriction $\mathcal{C}_{time}$ or non-monotonic progressions,  
yielding a valid subset $\widetilde{\mathcal{C}} \subseteq \mathcal{C}$ that satisfies all temporal constraints.  

\myparagraphunderline{Semantic–temporal re-ranking}
From $\widetilde{\mathcal{C}}$, candidates are re-ranked by jointly considering (i) semantic compatibility with the indicator and (ii) proximity of their timestamps to the expected reference time.  
Let $\textsf{DRM}(\Ind_i,p)$ denote the semantic similarity between the indicator and path, 
and $t(\cdot)$ denote the representative timestamp.  
Each candidate receives a composite score:
\[
\text{Score}(p)
= \lambda_{\text{sem}}\cdot \textsf{DRM}(\Ind_i,p)
+ \lambda_{\text{prox}}\cdot \exp \bigl(-|\,t(p)-t(\Ind_i)\,|/\sigma\bigr),
\]
where $\lambda_{\text{sem}}$ and $\lambda_{\text{prox}}$ balance semantic alignment and temporal proximity.  
The top-$W_1$ candidates are retained as a compact, high-quality pool
for LLM-based selection.

\myparagraphunderline{LLM-aware selection}
Following the re-ranking procedure, the candidate paths are reduced to $W_1$. We then prompt LLM to score and select the top-$W_{\max}$ reasoning paths most likely to satisfy question’s temporal and semantic constraints. 
This final step emphasizes faithfulness and interpretability, producing a minimal but high-precision evidence set for the sufficiency verification in the reasoning loop.

\subsection{Experience Memory}
\label{sec:memory}


After each reasoning episode, verified temporal paths and toolkit decisions are stored for future reuse, 
forming the foundation of the experience memory layer.  
This layer serves as the long-term knowledge base of \mot, 
recording reasoning trajectories, toolkit selection patterns, and verified temporal facts.  
It bridges different reasoning cycles by allowing the model to retrieve relevant exemplars 
and continually refine its decision-making through accumulated experience.  
Unlike static prompt libraries, the memory operates as a dynamic, self-evolving component supporting both retrieval and continual adaptation across diverse question types.
The pseudo-code of the section is shown in Algorithm \ref{algorithm:memory} of Appendix \ref{appendix:alg}.

\myparagraph{Experience retrieval}
During the Tree of Time reasoning process, each sub-task may query the experience pool 
$\mathcal{E}_{pool}$ to obtain relevant exemplars for decomposition, seed selection, or toolkit usage.  
All retrieval operations (e.g., $\text{GetTypeExp}$, $\text{GetDecompExp}$, $\text{GetToolkitExp}$) 
are implemented under a unified interface: 
$\mathcal{E} = 
\textsf{RetrieveExperience}(\mathcal{E}_{pool}, q_i, \Ind_i, \tau, W_{\text{exp}}),$
where $\tau=\text{Type}(Q)$ denotes the temporal type of the current reasoning context. 
Each query is restricted to the memory subset that shares the same type label $\tau$, 
ensuring contextual consistency and preventing cross-type noise.  
Every experience record stores both textual metadata and its dense embeddings, one for the question text 
and another for its indicator, allowing semantic–temporal similarity search through a FAISS-based index \cite{douze2024faiss}.  The incorrect samples are used as warnings.

To accelerate lookup, a high-frequency buffer caches recently accessed exemplars.  
Entries in this buffer are ranked by a hybrid metric combining embedding similarity and hit frequency:
$
\textsf{Score}(E_j) = \lambda_{\text{sim}}\cos(e_{q_i}, e_{E_j}) 
+ \lambda_{\text{hit}}\textsf{Count}(E_j),
$
so that frequently used exemplars gain higher reuse priority.  
This dual-layer design allows \mot\ to balance long-term generalization with short-term adaptability during ongoing reasoning.

\begin{table}[t]
\centering
\caption{Performance comparison on \textbf{MultiTQ} (Hits@1, \%). Best in \textbf{bold}, second-best \underline{underlined}.}
\vspace{-3mm}
\label{tab:main_results_multiTQ}
\resizebox{\columnwidth}{!}{
\begin{tabular}{l c c c c c c}
\toprule
\multirow{2}{*}{\textbf{Model}} & \multirow{2}{*}{\textbf{LLM}} & \multirow{2}{*}{\textbf{Overall}} & \multicolumn{2}{c}{\textbf{Question Type}} & \multicolumn{2}{c}{\textbf{Answer Type}} \\
\cmidrule(lr){4-5}\cmidrule(lr){6-7}
 &  &  & \textbf{Multiple} & \textbf{Single} & \textbf{Entity} & \textbf{Time} \\
\midrule
BERT\cite{devlin2019bert}        & \multirow{2}{*}{--} & 8.3 & 6.1 & 9.2 & 10.1 & 4.0 \\
ALBERT\cite{lan2019albert}      &                      & 10.8 & 8.6 & 11.6 & 13.9 & 3.2 \\
\midrule
EmbedKGQA\cite{saxena2020improving}   & \multirow{3}{*}{--} & 20.6 & 13.4 & 23.5 & 29.0 & 0.1 \\
CronKGQA\cite{saxena2021question}    &                      & 27.9 & 13.4 & 33.7 & 32.8 & 15.6 \\
MultiQA\cite{chen2023multi}     &                      & 29.3 & 15.9 & 34.7 & 34.9 & 15.7 \\
\midrule
ChatGPT     & \multirow{6}{*}{GPT-3.5-Turbo} & 10.2 & 7.7  & 14.7 & 13.7 & 2.0 \\
KG-RAG\cite{ARI}      &                                     & 18.5 & 16.0 & 20.0 & 23.0 & 7.0 \\
ReAct KB\cite{ARI}    &                                     & 21.0 & 13.6 & 63.5 & 31.3 & 30.0 \\
ARI\cite{ARI}         &                                     & 38.0 & 68.0 & 21.0 & 39.4 & 34.4 \\
TempAgent\cite{qianyihu-etal-2025-time}   &                                     & 53.9 & 16.8 & 68.4 & 47.8 & 66.1 \\
\midrule
\multirow{4}{*}{\textbf{MemoTime}} 
 & GPT-4o-mini  & 64.2 & 40.3 & 73.0 & 61.5 & 70.8 \\
 & Qwen3-32B    & 68.2 & 42.5 & 77.6 & 62.1 & 81.7 \\
 & DeepSeek-V3  & \underline{73.0} & \underline{45.9} & \underline{82.9} & \underline{67.7} & \underline{84.6} \\
 &\cellcolor{blue!6}  GPT-4-Turbo  & \cellcolor{blue!6} \textbf{77.9} & \cellcolor{blue!6} \textbf{53.8} & \cellcolor{blue!6} \textbf{86.8} & \cellcolor{blue!6} \textbf{74.5} & \cellcolor{blue!6} \textbf{85.3} \\
\bottomrule
\end{tabular}
}
\end{table}

\myparagraph{Experience update}
After a reasoning cycle concludes, all verified sub-questions, selected toolkits, and reasoning paths 
are recorded back into $\mathcal{E}_{pool}$.  
Each record includes textual content, structured indicators, execution parameters, temporal constraints, 
and corresponding embeddings for both the question and the indicator.  
This dual representation supports precise semantic alignment and fast vector retrieval.  
New entries are inserted into both the global memory and the buffer; 
obsolete or low-value traces are periodically pruned to prevent redundancy.  
Through continuous recording and pruning, the experience pool evolves alongside the system’s reasoning behaviour, 
serving as a compact yet expressive representation of accumulated knowledge.

\myparagraphunderline{Cross-type augmentation}
While retrieval is type-restricted, the memory layer supports cross-type augmentation during updates.  
When a sub-question from one temporal type exhibits high structural similarity with exemplars in another, 
the new record is annotated with multiple secondary type labels.  
This mechanism enables \mot\ to share transferable reasoning patterns across related temporal operators 
(\eg  \text{BeforeLast} and \text{AfterFirst}), forming an interconnected experience graph that 
improves recall and diversity without sacrificing retrieval precision.

\myparagraph{Continual adaptation}
The memory layer continuously re-weights and reorganizes stored exemplars based on retrieval frequency, 
temporal freshness, and reasoning success.  
High-confidence exemplars are prioritized in future retrievals, 
while outdated or inconsistent examples gradually decay in influence.  
The embedding index is periodically rebalanced to maintain uniform coverage and retrieval efficiency.  
This self-adaptive mechanism ensures that \mot\ focuses on temporally relevant reasoning strategies 
and generalizes effectively to unseen question types.

\myparagraphunderline{Integration with reasoning}
The experience memory interacts closely with the Tree of Time reasoning loop described in Section~\ref{sec:recursion}.  
At runtime, it acts as a shared repository for experience-guided decomposition, toolkit selection, and answer synthesis.  
After each reasoning session, verified results, including embeddings, type labels, sufficiency status, and hit statistics, are 
propagated back into memory, updating both the long-term index and the fast-access buffer.  
This design closes the reasoning–learning loop, enabling \mot\ to continually refine its temporal reasoning ability 
through iterative retrieval, execution, and self-improvement.

\vspace{-2mm}
\section{Experiments}
\label{sec:exp}
In this section, we evaluate \mot on two challenge TKGQA datasets. The detailed experimental settings, including datasets, baselines, and implementations, can be found in Appendix \ref{appen:dataset_details}.

\begin{table}[t]
\centering
\caption{Performance comparison on \textbf{TimeQuestions} (Hits@1, \%). Best in \textbf{bold}, second-best \underline{underlined}.}
\vspace{-3mm}
\label{tab:main_results_timeQ}
\resizebox{\columnwidth}{!}{
\begin{tabular}{l c c c c c c}
\toprule
\multirow{2}{*}{\textbf{Model}} & \multirow{2}{*}{\textbf{LLM}} & \multirow{2}{*}{\textbf{Overall}} & \multicolumn{2}{c}{\textbf{Question Type}} & \multicolumn{2}{c}{\textbf{Answer Type}} \\
\cmidrule(lr){4-5}\cmidrule(lr){6-7}
 &  &  & \textbf{Explicit} & \textbf{Implicit} & \textbf{Temporal} & \textbf{Ordinal} \\
\midrule
PullNet\cite{sun2019pullnet}   & \multirow{3}{*}{--} & 10.5 & 2.2  & 8.1  & 23.4 & 2.9 \\
Uniqorn\cite{pramanik2024uniqorn}   &                      & 33.1 & 31.8 & 31.6 & 39.2 & 20.2 \\
GRAFT-Net\cite{sun2018open} &                      & 45.2 & 44.5 & 42.8 & 51.5 & 32.2 \\
\midrule
CronKGQA\cite{mavromatis2022tempoqr}  & \multirow{5}{*}{--} & 46.2 & 46.6 & 44.5 & 51.1 & 36.9 \\
TempoQR\cite{mavromatis2022tempoqr}   &                      & 41.6 & 46.5 & 3.6  & 40.0 & 34.9 \\
EXAQT\cite{jia2021complex}     &                      & 57.2 & 56.8 & 51.2 & 64.2 & 42.0 \\
TwiRGCN\cite{sharma2022twirgcn}   &                      & 60.5 & 60.2 & 58.6 & 64.1 & 51.8 \\
LGQA\cite{liu2023local}      &                      & 52.9 & 53.2 & 50.6 & 60.5 & 40.2 \\
\midrule
GenTKGQA\cite{gao2024two}  & Fine-tune & 58.4 & 59.6 & \underline{61.1} & 56.3 & \underline{57.8} \\
TimeR$^4$\cite{T4} & Fine-tune & 64.8 & \underline{66.0} & 52.9 & \textbf{77.6} & 45.5 \\
ChatGPT   & GPT-3.5-Turbo & 45.9 & 43.3 & 51.1 & 46.5 & 48.1 \\
\midrule
\multirow{4}{*}{\textbf{MemoTime}}
 & GPT-4o-mini  & 60.9 & 60.0 & 50.0 & 69.4 & 52.8 \\
 & Qwen3-32B    & 60.6 & 58.3 & 51.1 & 69.8 & 53.3 \\
 & DeepSeek-V3  & \underline{67.8} & 63.1 & 53.3 & 71.3 & 54.3 \\
 &\cellcolor{blue!6}  GPT-4-Turbo  & \cellcolor{blue!6} \textbf{71.4} & \cellcolor{blue!6} \textbf{67.0} & \cellcolor{blue!6} \textbf{67.5} & \cellcolor{blue!6} \underline{74.5} & \cellcolor{blue!6} \textbf{58.7} \\
\bottomrule
\end{tabular}
}
\end{table}

\begin{table*}[t]
\centering
\caption{Performance comparison between the IO baseline and \mot across two temporal QA datasets using six backbone LLMs. 
The highest improvement is highlighted in bold, and the second-best is underlined for each data.}
\label{tab:backbone_memotime}
\vspace{-2mm}
\resizebox{1\textwidth}{!}{
\begin{tabular}{l
ccc ccc ccc ccc ccc ccc}
\toprule
\multirow{2}{*}{\textbf{Dataset}} &
\multicolumn{3}{c}{\textbf{Qwen3-4B}} &
\multicolumn{3}{c}{\textbf{Qwen3-8B}} &
\multicolumn{3}{c}{\textbf{Qwen3-32B}} &
\multicolumn{3}{c}{\textbf{Qwen3-80B}} &
\multicolumn{3}{c}{\textbf{DeepSeek-V3}} &
\multicolumn{3}{c}{\textbf{GPT-4-Turbo}} \\[2pt]
\cmidrule(lr){2-4}\cmidrule(lr){5-7}\cmidrule(lr){8-10}\cmidrule(lr){11-13}\cmidrule(lr){14-16}\cmidrule(lr){17-19}
&\textbf{ IO }& \textbf{\mot} & \textbf{$\uparrow$}
& \textbf{IO }& \textbf{\mot }& \textbf{$\uparrow$}
& \textbf{IO} & \textbf{\mot} & \textbf{$\uparrow$}
& \textbf{IO} & \textbf{\mot }& \textbf{$\uparrow$}
& \textbf{IO }& \textbf{\mot} & \textbf{$\uparrow$}
&\textbf{ IO }& \textbf{\mot }& \textbf{$\uparrow$} \\
\midrule
\textbf{MultiTQ }
& 3.5  & 55.3 & 14.8 $\times$ 
&\cellcolor{blue!6} 2.5  &\cellcolor{blue!6} {\ul 57.0} & \cellcolor{blue!6}{\ul 21.8 $\times$ }
& \cellcolor{blue!6}1.3  & \cellcolor{blue!6}\textbf{61.4} &\cellcolor{blue!6} \textbf{46.2 $\times$ }
& 9.5  & 67.5 & 6.1 $\times$ 
& 7.5  & 70.9 & 8.5 $\times$ 
& 12.0 & 76.3 & 5.4 $\times$ \\
\textbf{TimeQuestion}
& \cellcolor{blue!6}18.0 &\cellcolor{blue!6} \textbf{45.3} & \cellcolor{blue!6}\textbf{1.5 $\times$ }
& 28.5 & 49.4 & 0.7 $\times$ 
&\cellcolor{blue!6} 30.0 &\cellcolor{blue!6} {\ul 60.6} & \cellcolor{blue!6}{\ul1.0 $\times$ }
& 41.0 & 62.7 & 0.5 $\times$ 
& 43.5 & 64.6 & 0.5 $\times$ 
& 44.0 & 68.1 & \text{0.5 $\times$ } \\
\bottomrule
\end{tabular}}
\end{table*}

\subsection{Main results}  

We evaluate \mot against a comprehensive set of baselines on 
\text{MultiTQ} and \text{TimeQuestions}, as shown in Table~\ref{tab:main_results_multiTQ} and Table~\ref{tab:main_results_timeQ}. 
%
On the \text{MultiTQ} dataset, traditional pre-trained language models (BERT, ALBERT) and embedding-based methods (EmbedKGQA, CronKGQA, MultiQA) show limited temporal reasoning ability, with overall accuracies below 30\%.  
Their static embeddings fail to capture multi-hop or cross-entity temporal dependencies.  
LLM-based methods, e.g, KG-RAG, ReAct KB, ARI, TempAgent, leverage prompt reasoning but remain sensitive to implicit time expressions and struggle with multi-constraint temporal alignment.  
In contrast, MemoTime consistently outperforms all competitors across both question and answer types.  
MemoTime with GPT-4-Turbo achieves an overall 77.9\% Hit@1, surpassing TempAgent by 24.0\% and outperforming all GPT-3.5 baselines by a large margin.  
Even smaller backbones (e.g., Qwen3-32B) reach 68.2\%, exceeding the best GPT-3.5-based models.  
These results confirm that MemoTime effectively integrates temporal grounding, hierarchical reasoning, and operator-aware retrieval to ensure temporal faithfulness and reasoning stability.
On the \text{TimeQuestions} dataset, which emphasizes explicit versus implicit temporal expressions and temporal versus ordinal answer types, MemoTime again achieves the best overall results.  
As shown in Table~\ref{tab:main_results_timeQ}, MemoTime with GPT-4-Turbo achieves 71.4\% overall accuracy and 74.5\% on temporal-type questions, outperforming the fine-tuned GenTKGQA and TimeR$^4$ models, despite being fully training-free.  
This demonstrates MemoTime’s ability to handle implicit temporal relations and operator combinations through memory-guided decomposition and dynamic temporal retrieval.  
Overall, across both datasets, MemoTime achieves new state-of-the-art results under all backbone LLMs, validating its design as a \text{memory-augmented temporal reasoning framework} capable of learning from experience, synchronizing multiple entities, and maintaining temporal consistency across diverse question types.

\vspace{-2mm}

\subsection{Ablation Study}\label{exp:ablation}

\myparagraphquestion{How does the performance of \mot vary across different LLM backbones}
To evaluate the generality of our approach, we tested \mot on six LLM backbones, including Qwen3-4B, Qwen3-8B, Qwen3-32B, Qwen3-80B, DeepSeek-V3, and GPT-4-Turbo, across two temporal QA benchmarks (MultiTQ and TimeQuestions). 
As shown in Table~\ref{tab:backbone_memotime}, \mot consistently improves performance over the IO baseline across all settings. 
On the more complex \text{MultiTQ}, which requires multi-hop and cross-entity temporal reasoning, \mot boosts accuracy from as low as 1.3\% to 61.4\% on Qwen3-32B, achieving a \text{46.2 times} relative gain. 
Even the smallest backbones (e.g., Qwen3-4B) experience a \text{14.8 times} improvement, demonstrating that our recursive, memory-augmented reasoning compensates for weaker temporal understanding in smaller models.
For \text{TimeQuestions}, which involves simpler, single-hop temporal relations, \mot still improves performance modestly by up to 1.5 times, indicating consistent stability across reasoning difficulty levels.  
Overall, \mot enables smaller models to perform competitively with large-scale LLMs such as GPT-4-Turbo, while stronger models continue to benefit from enhanced temporal faithfulness and reduced reasoning variance. 
These results confirm that \mot generalizes well across architectures and scales, serving as a plug-in temporal reasoning module rather than relying on model-specific fine-tuning.

\begin{table}[t]
\centering
\caption{Ablation study on \textbf{MultiTQ} (Hits@1, \%).}
\vspace{-2mm}
\label{tab:ablation_memotime}
\resizebox{\columnwidth}{!}{
\begin{tabular}{lccccc}
\toprule
\multirow{2}{*}{\textbf{Model Variant}} & \multirow{2}{*}{\textbf{Overall}} & 
\multicolumn{2}{c}{\textbf{Question Type}} & 
\multicolumn{2}{c}{\textbf{Answer Type}} \\ 
\cmidrule(lr){3-4} \cmidrule(lr){5-6}
& & \textbf{Multiple} & \textbf{Single} & \textbf{Entity} & \textbf{Time} \\
\midrule
\textbf{MemoTime (Full w/GPT-4o-mini)}                & \textbf{64.2} & \textbf{40.3} & \textbf{73.0} & \textbf{61.5} & \textbf{70.8} \\
\textit{w/o Graph-based Retrieval}      & 52.9          & 23.5          & 65.0          & 48.2          & 64.2          \\
\textit{w/o Embedding Retrieval}        & 60.1          & 38.6          & 68.0          & 57.4          & 66.8          \\
\textit{w/o Temporal Evidence Retrieval}& 11.2          & 8.0           & 12.4          & 11.5          & 12.0          \\
\textit{w/o Question Tree}              & 58.3          & 19.3          & 72.6          & 54.6          & 70.1          \\
\textit{w/o Experience Memory}          & 59.8          & 26.1          & 72.2          & 55.8          & 69.6          \\
\bottomrule
\end{tabular}}
\end{table}

\myparagraphquestion{How does temporal evidence retrieval affect performance}
To assess the contribution of the temporal evidence retrieval module, we disable different retrieval components.
As shown in Table~\ref{tab:ablation_memotime}, removing graph-based retrieval reduces the overall accuracy from 64.2\% to 52.9\%, with a substantial drop on multiple-entity questions (from 40.3\% to 23.5\%), indicating that explicit structural traversal is essential for maintaining temporally aligned entity paths.
When embedding-based retrieval is removed, the performance decreases moderately from 64.2\% to 60.1\%, suggesting that semantic retrieval complements the graph search by capturing lexically diverse or contextually implicit temporal cues.
In contrast, when all external retrieval modules are disabled and the model relies solely on its internal knowledge, the accuracy falls drastically to 11.2\%, reflecting a near-complete loss of temporal grounding.
These findings confirm that temporal evidence retrieval is not only indispensable but also synergistic: graph-based retrieval ensures factual precision through topological consistency, while embedding-based retrieval enhances coverage through semantic generalization.

\myparagraphquestion{How does hierarchical decomposition affect performance} 
To examine the role of hierarchical decomposition, we disable the Tree-of-Time module, forcing the model to reason over the full question without recursive planning.
As shown in Table \ref{tab:ablation_memotime},
the overall accuracy increases from 64.2\% to 58.3\%, while performance on multiple-entity questions drops sharply from 40.3\% to 19.3\%.
This substantial decline demonstrates that recursive decomposition is critical for handling complex temporal dependencies and multi-hop constraints.
Even with retrieval retained, the absence of decomposition limits the model’s ability to enforce monotonic timestamp ordering and to synchronize temporal operators across entities, leading to degraded reasoning coherence.

\myparagraphquestion{How does experience memory affect performance} 
To analyse the role of continual learning, we remove the experience memory module, preventing \mot from accessing previously verified reasoning traces.
The result in Table \ref{tab:ablation_memotime} shows it causes a moderate performance drop from 64.2\% to 59.8\%, with larger reductions observed on multi-hop questions, from 40.3\% to 26.1\%.
The results indicate that experience memory enhances reasoning stability and efficiency.
Although the model remains competitive without it, the memory mechanism contributes to more consistent reasoning trajectories and progressive self-improvement across inference cycles.

To further evaluate the effectiveness and efficiency of \mot, we conduct additional experiments, including multi-granular temporal QA analysis, efficiency analysis, 
and case studies of temporal interpretability and
faithful reasoning
in Appendix~\ref{appendxi:add_exp}.

\vspace{-2mm}
\section{Conclusion}
\label{sec:conclusion}

In this paper, we present \textbf{MemoTime}, a memory-augmented temporal knowledge graph framework that enhances LLM temporal reasoning through structured grounding, hierarchical decomposition, and continual experience learning. 
MemoTime enables temporally faithful, multi-entity reasoning by constructing operator-aware temporal paths and dynamically reusing verified reasoning traces. 
Its adaptive retrieval and memory integration allow it to balance efficiency, accuracy, and temporal consistency across diverse reasoning scenarios. 
Extensive experiments on multiple temporal QA datasets demonstrate that \mot
outperforms existing baselines, showcasing its superior reasoning
capabilities and interoperability.

\begin{acks}
Xiaoyang Wang is supported by the Australian Research Council DP230101445 and DP240101322. Wenjie Zhang is supported by the Australian Research Council DP230101445 and FT210100303.
\end{acks}

\bibliographystyle{ACM-Reference-Format}
\bibliography{acmart}

\newcommand{\Dep}{\mathrm{D}}

\newpage
\appendix

\section{Algorithm}\label{appendix:alg}

\subsection{Temporal Grounding}\label{appendix:alg:analysis}
We present the comprehensive algorithmic procedure for temporal grounding (Section \ref{sec:method:grounding}) in Algorithm \ref{algorithm:analysis}.

\subsection{Tree of Time Reasoning}\label{appendix:alg:controller}
We present the comprehensive algorithmic procedure for
Tree of Time hierarchical reasoning (Section \ref{sec:recursion}) in Algorithm \ref{algorithm:controller}.

\subsection{Temporal Evidence Retrieval and Pruning}\label{appendix:alg:execution}
We present the comprehensive algorithmic procedure for temporal evidence retrieval and pruning (Section \ref{sec:retrieval}) in Algorithm \ref{algorithm:execution}.

\subsection{Experience Memory}\label{appendix:alg:memory}
We present the comprehensive algorithmic procedure for experience memory (Section \ref{sec:memory}) in Algorithm \ref{algorithm:memory}.

\begin{algorithm}[h]
{
{
\SetVline
\small
\caption{{\small{TemporalGrounding}}}\label{algorithm:analysis}

\Input{Question $Q$, TKG $\mathcal{G}$, experience pool $\mathcal{E}_{pool}$, limits $D_{\max}, W_{\exp}$}
\Output{Subquestion tree $\mathcal{T}_Q$, indicators $\{\Ind_i\}$, topic entities $Topics$, type $\textsf{Type}(Q)$, subgraph $\mathcal{G}_Q$}

\CmtState{\\$Keywords \leftarrow \textsf{NER}(Q)$; $Cand \leftarrow \textsf{LinkToKG}(Keywords,\mathcal{G})$}{LLM extraction + DRM alignment}
\State{$Topics \leftarrow \textsf{Disambiguate}(Q,Cand)$}

\CmtState{\\$\mathcal{G}_Q \leftarrow \textsf{BuildTemporalSubgraph}(Topics,\mathcal{G}, D_{\max})$}{bounded $D_{\max}$-hop temporal neighborhood}

\CmtState{\\$\mathcal{E}_{type} \leftarrow \textsf{GetTypeExp}(\mathcal{E}_{pool}, Q, W_{\exp})$}{Temporal type classification}
\State{$\textsf{Type}(Q) \leftarrow \textsf{Prompt}_{\text{TypeSelect}}(Q, \mathcal{E}_{type})$}

\State{\textbf{Return} $\mathcal{T}_Q,\{\Ind_i\}, Topics, \textsf{Type}(Q), \mathcal{G}_Q$}
}
}
\end{algorithm}
\begin{algorithm}[h]
{
{
\SetVline
\small
\caption{{\small{TreeofTimeReasoning}}}\label{algorithm:controller}

\Input{indicators $\{\Ind_i\}$, subgraph $\mathcal{G}_Q$, topics $Topics$, experience pool $\mathcal{E}_{pool}$, limits $D_{\max}, B_{\max}, W_{\max}, W_{\exp}$}
\Output{Local answers/paths per node; updated tree}


\State{$\mathcal{E}_{decomp} \leftarrow \textsf{GetDecompExp}(\mathcal{E}_{pool}, Q, W_{\exp}, \textsf{Type}(Q))$}
\If{$\exists$ compatible plan in $\mathcal{E}_{decomp}$}{
  \State{$\mathcal{T}_Q \leftarrow \textsf{ReusePlan}(\mathcal{E}_{decomp}, Q, Topics)$}
}
  \State{\textbf{else} $\mathcal{T}_Q \leftarrow \textsf{Prompt}_{\text{Decompose}}(Q, \mathcal{E}_{decomp}, \textsf{Type}(Q))$}

\State{$\{\Ind_i\} \leftarrow \textsf{ExtractIndicators}(\mathcal{T}_Q)$}
\ForEach{$(q_i,\Ind_i) \in \textsf{TraverseRootToLeaf}(\mathcal{T}_Q)$}{
  \State{$(\hat{a},\hat{P},\texttt{sufficient}) \leftarrow \textsf{MemoryLookupAndTest}(q_i,\Ind_i,\mathcal{E}_{pool})$}
  \State{\textbf{if} $\texttt{sufficient}$ \textbf{then} \textbf{continue}}

  \State{$\mathcal{E}_{tool} \leftarrow \textsf{GetToolkitExp}(\mathcal{E}_{pool}, q_i,\Ind_i, \textsf{Type}(Q), W_{\exp})$}
  \State{$\mathcal{T}_i \leftarrow \textsf{Prompt}_{\text{ToolkitSelect}}(\mathcal{T}_\theta, \Ind_i, q_i, \mathcal{E}_{tool})$}

  \State{$(C_i, S_i) \leftarrow \textsf{TemporalRetrieveAndPrune}(\mathcal{G}_Q, q_i,\Ind_i, Topics, \mathcal{T}_i, \mathcal{E}_{pool}, W_{\max})$}

  \State{$(a_i, P_i, \texttt{sufficient}) \leftarrow \textsf{DebateVoteAndTest}(q_i,\Ind_i,C_i)$}

  \If{$\neg \texttt{sufficient}$}{
    \If{$\textsf{DepthOK}~\land~\textsf{BranchOK}$}{
      \State{$\{q_{new}\},\{\Ind_{new}\} \leftarrow \textsf{RefineOrDecompose}(q_i,\Ind_i,C_i)$}
      \State{$\textsf{UpdateTree}(\mathcal{T}_Q,\{q_{new}\},\{\Ind_{new}\})$}
    }
  }
    \State{\textbf{else} $\textsf{WriteBackIfSuccess}(\mathcal{E}_{pool}, q_i,\Ind_i, a_i, P_i, \mathcal{T}_Q)$}
}
\State{\textbf{Return} \{$a_i$\}, \{$p_i$\}, $\mathcal{T}_Q$}
}
}
\end{algorithm}

\begin{algorithm}[h]
{
{
\SetVline
\small
\caption{{\small{TemporalRetrieveAndPrune}}}\label{algorithm:execution}

\Input{Subquestion $(q_i,\Ind_i)$, topics $Topics$, selected toolkits $\mathcal{T}_i$, experience pool $\mathcal{E}_{pool}$, subgraph $\mathcal{G}_Q$, limits $W_{\max}, W_{\exp}, D_{\max}$}
\Output{Pruned candidates $C_i$, selected seeds $S_i$}

\State{$\mathcal{E}_{seed} \leftarrow \textsf{GetSeedExp}(\mathcal{E}_{pool}, q_i,\Ind_i, W_{\exp})$}
\State{$S_i \leftarrow \textsf{Prompt}_{\text{SeedSelect}}(Topics,\Ind_i,q_i,\mathcal{E}_{seed})$; $C \leftarrow \emptyset$}

\ForEach{$T_\theta \in \mathcal{T}_i$ \textbf{in parallel}}{
  \State{$C \leftarrow C \cup \textsf{TreeBasedTemporalPathRetrieval}(\mathcal{G}_Q, T_\theta,\Ind_i,S_i,D_{\max})$}
  \State{$C \leftarrow C \cup \textsf{DenseEmbedRetrieve}(\Ind_i, T_\theta, \textsf{Docu}(\mathcal{G}_Q))$}
}

\State{$\widetilde{C} \leftarrow \{\,p \in C \mid p \vDash \mathcal{C}_{time}(\Ind_i) \land \textsf{Monotone}(p)\,\}$}

\ForEach{$p \in \widetilde{C}$}{
  \State{$s_{\text{sem}}(p) \leftarrow \textsf{SemanticDRM}(\Ind_i,p)$}
  \State{$s_{\text{prox}}(p) \leftarrow \exp(-|t(p)-t(\Ind_i)|/\sigma)$}
  \State{$Score(p) \leftarrow \lambda_{\text{sem}} s_{\text{sem}}(p) + \lambda_{\text{prox}} s_{\text{prox}}(p)$}
}
\State{$\widetilde{C} \leftarrow \textsf{SelectTopPathandSort}(\widetilde{C}, W_1)$}

\State{$C_i \leftarrow \textsf{Prompt}_{\text{SelectPath}}(\widetilde{C}, q_i,\Ind_i, W_{\max})$}

\State{\textbf{Return} $C_i, S_i$}

\vspace{2mm}
\textbf{Procedure} \textsf{TreeBasedTemporalPathRetrieval}$(\mathcal{G}_Q, T_\theta, \Ind_i, S_i, D_{\max})$\\
\SetVline
\small
\State{$\text{StartList}, \mathcal{C}_{time}, D_{\text{pred}} \leftarrow \textsf{ExtractInfo}(T_\theta,\Ind_i,S_i)$}
\State{$D \leftarrow 1$; $\text{Paths} \leftarrow \emptyset$; $\text{Frontier} \leftarrow \{(s, t_0, [])~|~s \in \text{StartList},~t_0=\textsf{InitTime}(\mathcal{C}_{time})\}$}
\While{$D \le D_{\max}$}{
  \State{$\text{NewFrontier} \leftarrow \emptyset$}
  \ForEach{$(v,t_{\text{last}},P) \in \text{Frontier}$}{
    \ForEach{$(v \xrightarrow{r,t_e} u) \in \textsf{ExpandOneHop}(\mathcal{G}_Q, v, T_\theta)$}{
      \If{$\neg \textsf{Monotone}(t_{\text{last}}, t_e)$}{\textbf{continue}}
      \State{$P' \leftarrow P \,\Vert\, (v \xrightarrow{r,t_e} u)$; $\text{Paths} \leftarrow \text{Paths} \cup \{P'\}$}
      \State{$\text{NewFrontier} \leftarrow \text{NewFrontier} \cup \{(u,t_e,P')\}$}
    }
  }
  \If{$|\text{NewFrontier}| > W_{\max}$}{\State{$\text{NewFrontier} \leftarrow \textsf{RelevantPruning}(\text{NewFrontier},\Ind_i,W_{\max})$}}
  \State{$\text{Frontier} \leftarrow \text{NewFrontier}$; $D \leftarrow D{+}1$}
}
\State{\textbf{Return} $\text{Paths}$}
}
}
\end{algorithm}

\begin{algorithm}
{
{
\SetVline
\small
\caption{{\small{Experience Memory}}}\label{algorithm:memory}

\Input{Experience pool $\mathcal{E}_{pool}$, query object (question or sub-question), indicator $\Ind$, type label $\tau$, buffer size $K$}
\Output{Matched exemplars $\mathcal{E}_{match}$ (optional), updated pool}

\State{$e_{\text{q}} \leftarrow \textsf{Encode}(q)$; $e_{\Ind} \leftarrow \textsf{Encode}(\Ind)$}
\State{$\mathcal{E}_{match} \leftarrow \textsf{ANN\_Search}(\mathcal{E}_{pool}, [e_{\text{q}},e_{\Ind}],~\text{filter: type}=\tau,~\text{top-}K)$}
\State{$\textsf{RankBy}(\lambda_{\text{sim}}\cdot\text{sim} + \lambda_{\text{hit}}\cdot\text{hit\_count})$}
\State{\textbf{Return} $\mathcal{E}_{match}$}

\vspace{1mm}
\textbf{Procedure} \textsf{WriteBackIfSuccess}$(\mathcal{E}_{pool}, q,\Ind,a,P)$\\
\SetVline
\small
\State{$\textsf{Store}(q,\Ind,a,P,\tau, \text{embeddings}=[e_{\text{q}},e_{\Ind}], \text{sufficient}=\texttt{true})$}
\State{$\textsf{UpdateBufferStats}(q,\Ind)$}

\vspace{1mm}
\textbf{Procedure} \textsf{MemoryLookupAndTest}$(q,\Ind,\mathcal{E}_{pool})$\\
\SetVline
\small
\State{$Hist \leftarrow \textsf{Retrieve}(q,\Ind, \text{filter: type}=\tau)$}
\ForEach{$(\hat{a},\hat{P}) \in Hist$}{
  \State{\textbf{if} $\textsf{Sufficient}(q,\Ind,\hat{a},\hat{P})$ \textbf{then} \textbf{Return} $\hat{a},\hat{P},\texttt{sufficient}=\texttt{true}$ }
}

\State{\textbf{Return} $\bot,\bot,\texttt{sufficient}=\texttt{false}$}
}
}
\end{algorithm}


\begin{table*}[h]
\centering
\caption{Multi-granularity temporal reasoning on Hits@1. Best in each column is in \textbf{bold}.}

\label{tab:multi_granularity}
\resizebox{0.8\linewidth}{!}{
\begin{tabular}{lccccccccc}
\toprule
\multirow{2}{*}{\textbf{Model}} &
\multicolumn{3}{c}{\textbf{Equal}} &
\multicolumn{3}{c}{\textbf{Before/After}} &
\multicolumn{3}{c}{\textbf{Equal-Multi}} \\
\cmidrule(lr){2-4}\cmidrule(lr){5-7}\cmidrule(lr){8-10}
& \textbf{Day} & \textbf{Month} & \textbf{Year}
& \textbf{Day} & \textbf{Month} & \textbf{Year}
& \textbf{Day} & \textbf{Month} & \textbf{Year} \\
\midrule
BERT            & 4.9  & 10.3 & 13.6 & 15.0 & 16.4 & 17.5 & 6.4  & 10.2 & 9.0  \\
DistillBERT     & 4.1  & 8.7  & 11.3 & 16.0 & 15.0 & 18.6 & 9.6  & 12.7 & 8.9  \\
ALBERT          & 6.9  & 8.2  & 13.2 & 22.1 & 27.7 & 30.8 & 10.3 & 14.4 & 14.4 \\
\midrule
EmbedKGQA       & 20.0 & 33.6 & 21.8 & 39.2 & 51.8 & 51.1 & 14.5 & 32.1 & 26.3 \\
CronKGQA        & 42.5 & 38.9 & 33.1 & 37.5 & 47.4 & 45.0 & 29.5 & 33.3 & 25.1 \\
MultiQA         & 44.5 & 39.3 & 35.0 & 37.9 & 54.8 & 52.5 & 30.8 & 32.1 & 28.3 \\
\midrule
\textbf{MemoTime w/ DeepSeek-V3}   & 85.6 & \textbf{85.1} & 92.3 & 81.4 & 77.4 & 82.4 & 60.0 & 55.3 & 53.2 \\
\textbf{MemoTime w/ GPT-4-Turbo}   & \textbf{85.8} & 85.0 & \textbf{93.8} & \textbf{92.5} & \textbf{86.2} & \textbf{94.1} & \textbf{68.4} & \textbf{70.4} & \textbf{64.4} \\
\bottomrule
\end{tabular}}
\end{table*}
\begin{table*}[h]
\caption{Efficiency analysis of \text{MemoTime} on \text{MultiTQ}.}
\label{tab:multitq_efficiency}
\centering
\resizebox{0.8\linewidth}{!}{
\begin{tabular}{lccccccc}
\toprule
\textbf{} & \textbf{Overall} & \textbf{After–First} & \textbf{Before–Last} & \textbf{Equal–Multi} & \textbf{First/Last} & \textbf{Before/After} & \textbf{Equal} \\
\midrule
\textbf{Average Depth }     & 1.37 & 2.12 & 2.00 & 1.57 & 1.00 & 1.54 & 1.04 \\
\textbf{Average Branch }    & 2.64 & 3.91 & 3.94 & 3.05 & 2.00 & 2.78 & 2.01 \\
\textbf{Average LLM Calls } & 8.75 & 11.55 & 11.49 & 9.88 & 7.42 & 9.10 & 7.32 \\
\textbf{Running Time (s)}   & 43.11 & 58.11 & 57.89 & 48.86 & 35.70 & 45.34 & 35.42 \\
\bottomrule
\end{tabular}}
\vspace{5mm}
\end{table*}

\vspace{5mm}
\section{Additional Experiments}
\label{appendxi:add_exp}

\subsection{Multi-Granular Temporal QA Analysis}
\label{appendxi:mutlti-}

We further evaluate \mot on temporal reasoning tasks that require distinguishing and aligning facts across different temporal granularities, including day, month, and year levels. 
This setting examines whether models can adapt to diverse temporal resolutions and maintain consistent reasoning performance.  
As shown in Table~\ref{tab:multi_granularity}, traditional pre-trained and embedding-based baselines exhibit significant degradation as the temporal resolution becomes finer or when multiple granularities are combined.
For example, while MultiQA achieves 44.5\% accuracy on the Equal–Day level, its performance drops sharply to 28.3\% on Equal–Multi–Year, indicating that these models struggle to maintain chronological consistency when reasoning across mixed time scales.

In contrast, \mot consistently achieves strong and stable results across all granularities and operators, reaching up to 94.1\% Hits@1 on Before/After–Year and maintaining 70.4\% under the most complex Equal–Multi–Month setting.
The consistent improvement across both fine-grained and compound temporal tasks demonstrates that \mot effectively adapts to the diversity of temporal reasoning requirements.  
This robustness is mainly attributed to two key components: (1) operator-aware decomposition, which dynamically adjusts reasoning depth and retrieval scope according to the temporal operator, and (2) temporal path construction, which enforces monotonic timestamp progression and co-aligns evidence across multiple resolutions.
Together, these mechanisms enable unified and faithful temporal reasoning while preserving interpretability across heterogeneous granularities.

\definecolor{HeadBG}{HTML}{F3F6FF}
\definecolor{RowAlt}{HTML}{FAFBFD}
\definecolor{BadgeBlue}{HTML}{E6F0FF}
\definecolor{BadgeGreen}{HTML}{E9F7EF}
\definecolor{BadgeOrange}{HTML}{FFF4E6}
\definecolor{BadgeGray}{HTML}{F2F2F2}

\newcommand{\badgeblue}[1]{\colorbox{BadgeBlue}{\strut\;\footnotesize #1\;}}
\newcommand{\badgegreen}[1]{\colorbox{BadgeGreen}{\strut\;\footnotesize #1\;}}
\newcommand{\badgeorange}[1]{\colorbox{BadgeOrange}{\strut\;\footnotesize #1\;}}
\newcommand{\badgegray}[1]{\colorbox{BadgeGray}{\strut\;\footnotesize #1\;}}

\setlength{\tabcolsep}{6pt}
\renewcommand{\arraystretch}{1.18}

\begin{table*}
\centering
\caption{MemoTime toolkit library with purposes, key parameters, and operator mappings.}
\label{tab:toolkits-color}
\resizebox{0.8\linewidth}{!}{%
\begin{tabular}{l l l l}
\rowcolor{HeadBG}
\toprule
\textbf{Toolkit} & \textbf{Purpose} & \textbf{Key parameters} & \textbf{Operator mapping} \\
\midrule
\rowcolor{RowAlt}
OneHop & One-hop neighbors with temporal filters 
& entity, direction, after/before, limit 
& \badgegray{seed} \;\badgeblue{local} \\
AfterFirst & First Nth event after a cutoff 
& entity, after, relation\_filter, limit=N 
& \badgegreen{after\_first} \\
\rowcolor{RowAlt}
BeforeLast & Last Nth event before a cutoff 
& entity, before, relation\_filter, limit=N 
& \badgeorange{before\_last} \\
BetweenRange & Events within a time window 
& entity, between=(start,end), granularity 
& \badgeblue{during}\; \badgeblue{between} \\
\rowcolor{RowAlt}
DayEvents & Global events on a specific date 
& date, relation\_filter, limit 
& \badgeblue{same-day}\; \badgegray{snapshot} \\
Month/YearEvents & Global events in a month/year 
& month or year, relation\_filter, limit 
& \badgeblue{same-month}\; \badgeblue{same-year} \\
\rowcolor{RowAlt}
DirectConnection & Direct edges between two entities 
& entity1, entity2, direction, time filters 
& \badgegray{pairwise}\; \badgeblue{validate} \\
Timeline & Chronological sequence for an entity 
& entity, direction, after/before, limit 
& \badgegreen{ordering}\; \badgeblue{stitch} \\
\bottomrule
\end{tabular}%
}
\vspace{5mm}
\end{table*}

\subsection{Efficiency Analysis}
\label{appendix:exp:efficiency}

To comprehensively evaluate the efficiency of \mot, we conduct three analyses on the MultiTQ dataset: LLM calls cost analysis, running time analysis, and computational cost analysis. 
Table~\ref{tab:multitq_efficiency} presents the averaged statistics across different question types, providing insights into reasoning efficiency and structural scalability.

\myparagraph{LLM calls cost analysis}
To measure the efficiency of utilizing LLMs, we analyze the average number of LLM calls required to complete a reasoning cycle across different question types. 
As shown in Table~\ref{tab:multitq_efficiency}, MemoTime performs an average of 8.75 calls per question, demonstrating its lightweight decomposition and reasoning pipeline. 
For simple operator types such as Equal and First/Last, the number of calls remains low (7–7.5), corresponding to a single reasoning round without recursive decomposition. 
In contrast, complex composite operators including After–First and Before–Last require deeper temporal validation, resulting in approximately 11.5 LLM calls. 
These results demonstrate that MemoTime maintains efficient reasoning, even for multi-hop and temporally constrained queries, by avoiding redundant invocations through structured decomposition and memory reuse.

\myparagraph{Running time analysis}
To further assess inference efficiency, we examine the average running time per question for different temporal types. 
The overall average runtime of MemoTime is 43.11 seconds per question, indicating an effective balance between retrieval and reasoning processes. 
Operator-specific analysis shows that single-hop or non-nested operators such as Equal and First/Last complete within approximately 35 seconds, whereas multi-hop operators like After–First and Before–Last require additional retrieval and temporal alignment, increasing the runtime to around 58 seconds. 
Despite these additional steps, MemoTime preserves scalability by dynamically pruning irrelevant temporal paths and parallelizing evidence collection within each hierarchical layer.

\myparagraph{Computational cost analysis}
To evaluate the structural efficiency of temporal reasoning, we analyze the average recursion depth and branching factor, which jointly characterize the complexity of decomposition and search. 
MemoTime exhibits a compact reasoning structure, with an average recursion depth of 1.37 and branching factor of 2.64, indicating efficient but expressive reasoning behavior. 
As shown in Table~\ref{tab:multitq_efficiency}, composite operators such as After–First and Before–Last require deeper reasoning hierarchies (depth around 2) and broader exploration (branch factor close to 4), whereas simpler operators such as Equal and First/Last remain shallow and nearly linear (depth near 1, branch factor around 2). 
These results demonstrate that MemoTime dynamically adjusts its computational footprint according to the temporal operator’s semantic complexity, achieving consistent temporal alignment while maintaining efficient reasoning performance.

\subsection{Case Study: Temporal Interpretability and Faithful Reasoning}
\label{sec:casestudy-temporal}
To demonstrate how \mot performs interpretable and temporally faithful reasoning,
in this section, we present case studies across different temporal question types in Tables~\ref{tab:case_after_first_final}--\ref{tab:case_between_final}.
Each case study illustrates how complex temporal questions are decomposed into sub-questions with structured indicators, toolkit selections, and retrieved factual quadruples. 
Through examples involving diverse operators such as \text{afterNfirst}, \text{beforeNlast}, and \text{between}, we show how \mot incrementally constructs temporal reasoning chains that maintain monotonic time ordering, respect operator-specific constraints, and yield consistent answers. 
These examples highlight how the proposed framework ensures transparency and temporal faithfulness throughout the reasoning process, producing clear, explainable evidence trails that align with human-understandable logical steps.

\section{Toolkit Library}
\label{sec:toolkit}

We implement eight specialized temporal retrieval toolkits exposing a unified interface with normalized parameters.
Table \ref{tab:toolkits-color} outlines their core purposes, key parameters, and corresponding temporal operators. 
The tools are composable and operator-aware under temporal constraints. They cover the majority of temporal reasoning patterns required for multi-hop question answering.

\section{Experiment Details}
\label{appen:dataset_details}

\myparagraph{Experiment Datasets}
\label{exp:Experiment datasets}
Previous studies have revealed that the widely used \text{CronQuestions} dataset~\cite{minMultihopReadingComprehension2019} contains spurious correlations that models can exploit to achieve inflated accuracy~\cite{sharma2022twirgcn}. 
Following the analysis in~\cite{T4}, we therefore evaluate \mot on two more reliable and temporally challenging benchmarks: \text{MultiTQ}~\cite{TKGQA} and \text{TimeQuestions}~\cite{jia2021complex}. 
\text{MultiTQ} is the largest publicly available Temporal Knowledge Graph Question Answering (TKGQA) dataset, constructed from the ICEWS05–15 event database, which records politically and socially significant events with precise timestamps. 
It consists of more than 500K question–answer pairs, covering 15 years of temporal scope. 
Each question is automatically generated from event triples $(h, r, t, \tau)$ and further refined to ensure diversity in natural language phrasing. 
The dataset supports multiple temporal granularities (year, month, and day), with questions distributed across over 3,600 calendar days. 
\text{MultiTQ} encompasses a wide range of temporal reasoning operators, including \text{before}, \text{after}, \text{equal}, \text{first}, and \text{last}, and contains both \text{single-hop} and \text{multi-hop} reasoning cases that require combining multiple temporal relations to derive the final answer.
In contrast, \text{TimeQuestions} is a smaller but linguistically richer benchmark designed to test compositional and relational temporal reasoning. 
It contains around 16K manually curated questions spanning four reasoning categories: \text{Explicit}, \text{Implicit}, \text{Temporal}, and \text{Ordinal}. 
Unlike \text{MultiTQ}, it uses only yearly granularity and focuses on questions that involve reasoning over alignment and temporal order rather than precise timestamps. 
Each question is grounded in Wikidata events and entities, with linguistic templates textasizing contextual reasoning, event sequencing, and temporal comparisons across multiple entities.
The detailed statistics of both datasets, including splits by reasoning type and dataset size, are summarized in Table~\ref{tab:datasets_stats_compact_2col}.


\myparagraph{Baselines}
We compare \mot against a comprehensive set of baselines covering three major methodological categories: pre-trained language models, knowledge graph embedding methods, and LLM-based reasoning methods. 

\text{On \text{MultiTQ}}, we include:
(1) Pre-trained language models (PLMs) such as BERT~\cite{devlin2019bert} and ALBERT~\cite{lan2019albert}, which are fine-tuned for QA over temporal contexts;
(2) Embedding-based temporal KGQA methods, including EmbedKGQA~\cite{saxena2020improving}, CronKGQA~\cite{saxena2021question}, and MultiQA~\cite{chen2023multi}, which leverage temporal embeddings to align question representations with time-stamped entity triples;
(3) LLM-based in-context learning (ICL) approaches, such as direct instruction prompting (IO), KG-RAG~\cite{ARI}, ReAct-KB~\cite{yao2022react,ARI}, ARI~\cite{ARI}, and TempAgent~\cite{qianyihu-etal-2025-time}, which combine TKG and reasoning via ChatGPT or similar models.

\text{On \text{TimeQuestions}}, we evaluate:
(1) Static KGQA baselines, including PullNet~\cite{sun2019pullnet}, Uniqorn~\cite{pramanik2024uniqorn}, and GRAFT-Net~\cite{sun2018open}, which operate on non-temporal knowledge graphs;
(2) Temporal KGQA baselines, including CronKGQA~\cite{saxena2021question}, TempoQR~\cite{mavromatis2022tempoqr}, EXAQT~\cite{jia2021complex}, and LGQA~\cite{liu2023local};
(3) LLM-based methods, including instruction prompting (IO) and two fine-tuned reasoning models: GenTKGQA~\cite{gao2024two}, which fine-tunes TKG with LLaMA2-7B, and TimeR$^4$~\cite{T4}, which couples a jointly fine-tuned retriever with ChatGPT reasoning.

For each baseline, we adopt the reported performance values from their original publications to ensure comparability under consistent evaluation settings. 
Following~\cite{tog1.0sun2023think, tog2.0ma2024think, pogtan2025paths}, we use \text{exact match accuracy} (Hits@1) as the principal evaluation metric. 
Recall and F1 scores are not used since knowledge sources are not limited to
document databases \cite{tog1.0sun2023think, tog2.0ma2024think}.

\myparagraph{Experiment Implementation}
All experiments are conducted using \text{GPT-4-Turbo} as the primary reasoning backbone for \mot. 
To validate the generality and plug-and-play adaptability of our framework, we further instantiate \mot with several alternative LLMs, including \text{DeepSeek-V3}, Qwen3-Next-80B-A3B-Instruct (\text{Qwen3-80B}), \text{Qwen3-32B}, \text{Qwen3-8B}, and \text{Qwen3-4B}. 
These models represent a diverse spectrum of capacities and architectural scales, allowing us to evaluate how \mot performs under different model sizes and decoding behaviors.
Following prior work~\cite{pogtan2025paths, tog1.0sun2023think}, the temperature is set to $0.4$ during the evidence exploration phase to encourage reasoning diversity, and to $0$ during the final answer generation stage to ensure deterministic output.The maximum generation length is 256 tokens.
Sentence-BERT \cite{devlin2019bert} is utilized as dense retrieval module (DRM).
For temporal path construction, we set $W_{\max}=3$, $D_{\max}=3$, $W_1=80$, and $\lambda_{\mathrm{sem}}=0.6$.
During the experience memory phase, we use $\lambda_{\mathrm{sim}}=0.6$, $\lambda_{\mathrm{hit}}=0.4$, and $W_{\mathrm{exp}}=10$.
All experiences are dynamically learned from verified reasoning traces, with five predefined exemplars used for cold-start initialization.
The database buffer size is set to 200 for efficiency.
The complete code implementation of \mot is publicly available.\footnote{\url{https://github.com/SteveTANTAN/Memotime}.}

\vspace{5mm}
\begin{table}[h]
\centering
\caption{Statistics of \text{MultiTQ} and \text{TimeQuestions}.}
\label{tab:datasets_stats_compact_2col}
\resizebox{0.8\linewidth}{!}{
\begin{tabular}{l l r r r}
\toprule
\multicolumn{2}{l}{\textbf{Category}} & \textbf{Train} & \textbf{Dev} & \textbf{Test} \\
\midrule
\multicolumn{5}{c}{\textbf{MultiTQ}} \\
\midrule
\multirow{3}{*}{\text{Single}}
  & Equal        & 135{,}890 & 18{,}983 & 17{,}311 \\
  & Before/After & 75{,}340  & 11{,}655 & 11{,}073 \\
  & First/Last   & 72{,}252  & 11{,}097 & 10{,}480 \\
\cmidrule(l){1-5}
\multirow{3}{*}{\text{Multiple}}
  & Equal-Multi  & 16{,}893 & 3{,}213 & 3{,}207 \\
  & After-First  & 43{,}305 & 6{,}499 & 6{,}266 \\
  & Before-Last  & 43{,}107 & 6{,}532 & 6{,}247 \\
\cmidrule(l){1-5}
\multicolumn{2}{l}{\textbf{Total (MultiTQ)}} & \textbf{386{,}787} & \textbf{57{,}979} & \textbf{54{,}584} \\
\midrule
\multicolumn{5}{c}{\textbf{TimeQuestions}} \\
\midrule
\multicolumn{2}{l}{Explicit} & 2{,}724 & 1{,}302 & 1{,}311 \\
\multicolumn{2}{l}{Implicit} & 651     & 291     & 292     \\
\multicolumn{2}{l}{Temporal} & 2{,}657 & 1{,}073 & 1{,}067 \\
\multicolumn{2}{l}{Ordinal}  & 938     & 570     & 567     \\
\cmidrule(l){1-5}
\multicolumn{2}{r}{\textbf{Total (TimeQuestions)}} & \textbf{6{,}970} & \textbf{3{,}236} & \textbf{3{,}237} \\
\bottomrule
\end{tabular}}
\end{table}

\begin{table*}[h]
\centering
\caption{Case study of  interpretability
and temporal faithfulness reasoning for “After the 2008 Olympics, which country was the first to sign an environmental treaty with \textbf{China}?”}
\label{tab:case_after_first_final}
\resizebox{1\linewidth}{!}{
\begin{tabularx}{0.97\textwidth}{@{}p{4.3cm}X@{}}
\toprule
\textbf{Question} & After the \textbf{2008 Olympics}, which country was the \textbf{first} to sign an environmental treaty with China? \\
\textbf{Temporal Type} & \textbf{afterNfirst} (requires $t_2>t_1$ and $\min(t_2)$) \\
\textbf{Topic Entities} & [Olympics 2008, China] \\
\toprule
\textbf{Overall Indicator} &
\begin{minipage}[t]{\linewidth}
Edges: $(\text{\textbf{Olympics 2008}},\,\text{opening},\,?x,\,t_1)$, $(?y,\,\text{sign environmental treaty},\,\text{\textbf{China}},\,t_2)$ \\
Constraints: $t_2>t_1$, $\text{after\_first}(t_2,t_1)$ \\
Time vars: $[t_1, t_2]$\vspace{2pt}
\end{minipage} \\
\toprule
\textbf{Q1} & \textbf{When was the 2008 Olympics held (opening anchor)?} \\
\textbf{Selected Seed} & [Olympics 2008] \\
\textbf{Indicator (quadruple)} & $(\text{Olympics 2008},\,\text{opening},\,?x,\,t_1)$ \\
\textbf{Time vars} & $[\text{same\_year}(t_1, 2008)]$ \\
\textbf{Toolkit \& Params} & \texttt{OneHop}(entity = Olympics 2008, after = 2008-01-01, before = 2009-01-01) \\
\textbf{Retrieved Facts} & 
\textbf{$(\text{Olympics 2008},\,\text{opening},\,\text{Beijing},\,\mathbf{2008\!-\!08\!-\!08})$};\\
& $(\text{Olympics 2008},\,\text{closing},\,\text{Beijing},\,2008\!-\!08\!-\!24)$ \\
\textbf{Sub-answer} & $t_1 = \mathbf{2008\!-\!08\!-\!08}$ (use opening as temporal anchor; ?x = Beijing) \\
\midrule
\textbf{Q2} & \textbf{After $\mathbf{2008\!-\!08\!-\!08}$, which country first signed an environmental treaty with China?} \\
\textbf{Selected Seed} & [China] \\
\textbf{Indicator (quadruple)} & $(?y,\,\text{sign environmental treaty},\,\text{China},\,t_2)$ \\
\textbf{Time vars} & $[\text{after}(t_2,\text{2008-08-08}), \text{after\_first}(t_2,\text{2008-08-08})]$ \\
\textbf{Toolkit \& Params} & \texttt{AfterFirst}(entity = China, after = \text{2008-08-08}, relation\_filter = sign environmental treaty) \\
\textbf{Retrieved Facts} &
\textbf{$(\text{Japan},\,\text{sign\_treaty\_env},\,\text{China},\,\mathbf{2009\!-\!02\!-\!10})$}; \\
& $(\text{Korea},\,\text{sign\_treaty\_env},\,\text{China},\,2009\!-\!07\!-\!18)$; \\
& $(\text{Germany},\,\text{sign\_treaty\_env},\,\text{China},\,2010\!-\!03\!-\!02)$ \\
\textbf{Sub-answer} & \textbf{Japan}, $t_2 = 2009\!-\!02\!-\!10$ (earliest valid time after \text{2008-08-08}) \\
\toprule
\textbf{Temporal Reasoning Chain} &
\textbf{Olympics 2008} $\xrightarrow[\text{2008-08-08}]{\text{opening}}$ Beijing 
$\;\;\leadsto\;\;$ \textbf{Japan} $\xrightarrow[\text{2009-02-10}]{\text{sign\_treaty\_env}}$ \textbf{China} (\textbf{after\_first} \text{2008-08-08}). \\
\textbf{Temporal Faithfulness} & All facts satisfy $t_2>t_1$; $\min(t_2)=2009$-02-10; monotonic ordering verified. \\
\textbf{Final Answer} & \textbf{Japan}. \\
\textbf{Response} &
“Anchoring $t_1$ at 2008-08-08 (opening of the Olympics), the earliest post-anchor treaty signing with China is by \textbf{Japan} on 2009-02-10, earlier than Korea (2009-07-18) and Germany (2010-03-02). Hence, the \text{after first} constraint selects Japan.” \\
\bottomrule
\end{tabularx}}
\end{table*}

\begin{table*}[t]
\centering
\caption{Case study of interpretability
and temporal faithfulness reasoning for “Who was the last leader to visit \textbf{Beijing} before 2010?”}
\label{tab:case_before_last_final}
\begin{tabularx}{0.97\textwidth}{@{}p{4.3cm}X@{}}
\toprule
\textbf{Question} & Who was the last leader to visit \textbf{Beijing} before 2010? \\
\textbf{Temporal Type} & \textbf{beforeNlast} (requires $t_1<2010$ and $\max(t_1)$) \\
\textbf{Topic Entities} & [Beijing, leader] \\
\toprule
\textbf{Overall Indicator} &
\begin{minipage}[t]{\linewidth}
Edges: $(?x,\,\text{visit},\,\text{\textbf{Beijing}},\,t_1)$ \\
Constraints: $t_1<2010$, $\text{before\_last}(t_1, 2010)$ \\
Time vars: $[t_1]$\vspace{2pt}
\end{minipage} \\
\toprule
\textbf{Q1} & \textbf{Which leaders visited Beijing before 2010?} \\
\textbf{Selected Seed} & [Beijing] \\
\textbf{Indicator (quadruple)} & $(?x,\,\text{visit},\,\text{Beijing},\,t_1)$ \\
\textbf{Time vars} & $[t_1<2010]$ \\
\textbf{Toolkit \& Params} & \texttt{Before}(entity = Beijing, before = 2010-01-01, relation\_filter = visit) \\
\textbf{Retrieved Facts} &
$(\text{Barack Obama},\,\text{visit},\,\text{Beijing},\,{2009\!-\!11\!-\!15})$; \\
& 
$(\text{Gordon Brown},\,\text{visit},\,\text{Beijing},\,2009\!-\!08\!-\!10)$; \\
& 
$(\text{Angela Merkel},\,\text{visit},\,\text{Beijing},\,2008\!-\!12\!-\!05)$ \\
\textbf{Sub-answer} & Candidate set = [Obama 2009-11-15, Brown 2009-08-10, Merkel 2008-12-05]. \\
\midrule
\textbf{Q2} & \textbf{Among them, who visited most recently before 2010?} \\
\textbf{Selected Seed} & [Candidate leaders] \\
\textbf{Indicator (quadruple)} & $(?x,\,\text{visit},\,\text{Beijing},\,t_1)$ \\
\textbf{Time vars} & $[\text{before}(t_1,2010),\;\text{last}(t_1)]$ \\
\textbf{Toolkit \& Params} & \texttt{FirstLast}(mode = last, relation\_filter = visit, before = 2010-01-01, sort = desc, limit = 1) \\
\textbf{Retrieved Facts} &
\textbf{$(\mathbf{Barack\,Obama},\,\text{visit},\,\text{Beijing},\,\mathbf{2009\!-\!11\!-\!15})$} \\
\textbf{Sub-answer} & \textbf{Barack Obama}, $t_1 = 2009$-11-15 (latest valid before 2010). \\
\toprule
\textbf{Temporal Reasoning Chain} &
\textbf{Barack Obama} $\xrightarrow[\text{2009-11-15}]{\text{visit}}$ \textbf{Beijing} (\textbf{before\_last} 2010). \\
\textbf{Temporal Faithfulness} & All events satisfy $t_1<2010$; $\max(t_1)=2009$-11-15; temporal order monotonic. \\
\textbf{Final Answer} & \textbf{Barack Obama}. \\
\textbf{Response} &
“Filtering visits to Beijing with $t_1<2010$ yields three leaders; the most recent timestamp (2009-11-15) corresponds to \textbf{Barack Obama}, fulfilling the \text{before last} constraint.” \\
\bottomrule
\end{tabularx}
\end{table*}

\newpage
\begin{table*}[t]
\centering
\caption{Case study of interpretability
and temporal faithfulness reasoning for “Before the 2010 Summit, which leader visited Beijing last?”}
\label{tab:case_before_last_final}
\begin{tabularx}{0.97\textwidth}{@{}p{4.3cm}X@{}}
\toprule
\textbf{Question} & Before the \textbf{2010 Summit}, which leader visited \textbf{Beijing} last? \\
\textbf{Temporal Type} & \textbf{beforeNlast} (requires $t_2<t_1$ and $\max(t_2)$) \\
\textbf{Topic Entities} & [2010 Summit, Beijing] \\
\toprule
\textbf{Overall Indicator} &
\begin{minipage}[t]{\linewidth}
Edges: $(\text{\textbf{Leader}},\,\text{visit},\,\text{\textbf{Beijing}},\,t_2)$, $(\text{\textbf{2010 Summit}},\,\text{held\_in},\,?x,\,t_1)$ \\
Constraints: $t_2<t_1$, $\text{before\_last}(t_2,t_1)$ \\
Time vars: $[t_1, t_2]$\vspace{2pt}
\end{minipage} \\
\toprule
\textbf{Q1} & \textbf{When was the 2010 Summit held?} \\
\textbf{Selected Seed} & [2010 Summit] \\
\textbf{Indicator (quadruple)} & $(\text{2010 Summit},\,\text{held\_in},\,?x,\,t_1)$ \\
\textbf{Time vars} & $[\text{specific\_year}(t_1, 2010)]$ \\
\textbf{Toolkit \& Params} & \texttt{OneHop}(entity = 2010 Summit, after = 2010-01-01, before = 2011-01-01) \\
\textbf{Retrieved Facts} & 
\textbf{$(\text{2010 Summit},\,\text{held\_in},\,\text{Toronto},\,\textbf{2010-06-26})$} \\
\textbf{Sub-answer} & $t_1 = \textbf{2010-06-26}$ (used as upper temporal boundary) \\
\midrule
\textbf{Q2} & \textbf{Before \text{2010-06-26}, which leader visited Beijing last?} \\
\textbf{Selected Seed} & [Beijing] \\
\textbf{Indicator (quadruple)} & $(?y,\,\text{visit},\,\text{Beijing},\,t_2)$ \\
\textbf{Time vars} & $[\text{before}(t_2,\text{2010-06-26}), \text{before\_last}(t_2,\text{2010-06-26})]$ \\
\textbf{Toolkit \& Params} & \texttt{BeforeLast}(entity = Beijing, before = \text{2010-06-26}, relation\_filter = visit) \\
\textbf{Retrieved Facts} &
\textbf{$(\textbf{Barack Obama},\,\text{visit},\,\text{Beijing},\,\textbf{2009-11-15})$}; \\
& $(\text{Gordon Brown},\,\text{visit},\,\text{Beijing},\,\text{2009-08-10})$; \\
& $(\text{Angela Merkel},\,\text{visit},\,\text{Beijing},\,\text{2008-12-05})$ \\
\textbf{Sub-answer} & \textbf{Barack Obama}, $t_2 = \text{2009-11-15}$ (latest valid time before \text{2010-06-26}) \\
\toprule
\textbf{Temporal Reasoning Chain} &
\textbf{Barack Obama} $\xrightarrow[\text{2009-11-15}]{\text{visit}}$ \textbf{Beijing} 
$\;\;\leadsto\;\;$ \textbf{2010 Summit} $\xrightarrow[\text{2010-06-26}]{\text{held\_in}}$ Toronto 
(\textbf{before\_last} \text{2010-06-26}). \\
\textbf{Temporal Faithfulness} & All facts satisfy $t_2<t_1$; $\max(t_2)=2009-11-15$; monotonic order verified. \\
\textbf{Final Answer} & \textbf{Barack Obama}. \\
\textbf{Response} &
“Anchoring $t_1$ at 2010-06-26 (2010 Summit), the last recorded visit to Beijing before this date was by \textbf{Barack Obama} on 2009-11-15. Earlier visits (Gordon Brown 2009-08-10, Angela Merkel 2008-12-05) confirm Obama as the before\_last case.” \\
\bottomrule
\end{tabularx}
\end{table*}
\newpage

\begin{table*}[t]
\centering
\caption{Case study of interpretability
and temporal faithfulness reasoning for “How many times did the \textbf{UN} hold a climate summit before 2020?”}
\label{tab:case_count_final}
\begin{tabularx}{0.97\textwidth}{@{}p{4.3cm}X@{}}
\toprule
\textbf{Question} & How many times did the \textbf{UN} hold a climate summit before 2020? \\
\textbf{Temporal Type} & \textbf{count + before} (requires $t_1<2020$ and aggregation over events) \\
\textbf{Topic Entities} & [UN, climate summit] \\
\toprule
\textbf{Overall Indicator} &
\begin{minipage}[t]{\linewidth}
Edges: $(\text{\textbf{UN}},\,\text{hold},\,\text{Summit},\,t_1)$ \\
Constraints: $t_1<2020$, $\text{topic}(\text{Summit})=\text{climate}$ \\
Time vars: $[t_1]$\vspace{2pt}
\end{minipage} \\
\toprule
\textbf{Q1} & \textbf{Which climate summits were held by the UN before 2020?} \\
\textbf{Selected Seed} & [UN] \\
\textbf{Indicator (quadruple)} & $(\text{UN},\,\text{hold},\,\text{Summit},\,t_1)$ \\
\textbf{Time vars} & $[t_1<2020]$ \\
\textbf{Toolkit \& Params} & \texttt{Before}(entity = UN, before = 2020-01-01, relation\_filter = hold, keyword = climate) \\
\textbf{Retrieved Facts} &
$(\text{UN},\,\text{hold},\,\text{ClimateSummit09},\,\mathbf{2009\!-\!12\!-\!07})$; \\
& 
$(\text{UN},\,\text{hold},\,\text{ClimateSummit13},\,\mathbf{2013\!-\!11\!-\!11})$; \\
& 
$(\text{UN},\,\text{hold},\,\text{ClimateSummit15},\,\mathbf{2015\!-\!12\!-\!05})$; \\
& 
$(\text{UN},\,\text{hold},\,\text{ClimateSummit19},\,\mathbf{2019\!-\!09\!-\!21})$; \\
& 
$(\text{UN},\,\text{hold},\,\text{ClimateSummit21},\,2021\!-\!11\!-\!01)$ \\
\textbf{Sub-answer} & Valid events before 2020 = 4 (2009, 2013, 2015, 2019). \\
\midrule
\textbf{Q2} & \textbf{Count all valid events that occurred before 2020.} \\
\textbf{Selected Seed} & [Filtered summits from Q1] \\
\textbf{Indicator (quadruple)} & $(\text{UN},\,\text{hold},\,\text{ClimateSummit\_i},\,t_1)$ \\
\textbf{Time vars} & $[\text{count}(t_1<2020)]$ \\
\textbf{Toolkit \& Params} & \texttt{Count}(filter = $t_1<2020$) \\
\textbf{Retrieved Facts} & Total = 4 events before 2020. \\
\textbf{Sub-answer} & \textbf{4 climate summits held before 2020.} \\
\toprule
\textbf{Temporal Reasoning Chain} &
\textbf{UN} $\xrightarrow[\text{2009–2019}]{\text{hold}}$ ClimateSummit\_i (\textbf{4 instances before 2020}). \\
\textbf{Temporal Faithfulness} & All timestamps $t_1<2020$; aggregation respects strict temporal filter. \\
\textbf{Final Answer} & \textbf{4 climate summits.} \\
\textbf{Response} &
“Filtering UN-held climate summits before 2020 yields four valid temporal events (2009, 2013, 2015, 2019).  
All satisfy $t_1<2020$, ensuring temporally faithful counting.” \\
\bottomrule
\end{tabularx}
\end{table*}
\newpage

\begin{table*}[t]
\centering
\caption{Case study of interpretability
and temporal faithfulness reasoning for “Between the 2015 Conference and the 2018 Summit, which company collaborated with Microsoft?”}
\label{tab:case_between_final}
\begin{tabularx}{0.97\textwidth}{@{}p{4.3cm}X@{}}
\toprule
\textbf{Question} & Between the \textbf{2015 Conference} and the \textbf{2018 Summit}, which company collaborated with \textbf{Microsoft}? \\
\textbf{Temporal Type} & \textbf{between} (requires $t_1 < t_3 < t_2$ and bounded interval) \\
\textbf{Topic Entities} & [2015 Conference, 2018 Summit, Microsoft] \\
\toprule
\textbf{Overall Indicator} &
\begin{minipage}[t]{\linewidth}
Edges: $(\text{\textbf{2015 Conference}},\,\text{held in},\,?x,\,t_1)$, $(\text{\textbf{2018 Summit}},\,\text{held in},\,?y,\,t_2)$, $(?z,\, \\ \text{collaborate with},\,\text{\textbf{Microsoft}},\,t_3)$ \\
Constraints: $t_1 < t_3 < t_2$, $\text{between}(t_3,[t_1,t_2])$ \\
Time vars: $[t_1, t_2, t_3]$\vspace{2pt}
\end{minipage} \\
\toprule
\textbf{Q1} & \textbf{When was the 2015 Conference held?} \\
\textbf{Selected Seed} & [2015 Conference] \\
\textbf{Indicator} & $(\text{2015 Conference},\,\text{held in},\,?x,\,t_1)$ \\
\textbf{Toolkit \& Params} & \texttt{OneHop}(entity = 2015 Conference, after = 2015-01-01, before = 2016-01-01) \\
\textbf{Retrieved Facts} & 
\textbf
{$(\text{2015 Conference},\,\text{held\_in},\,\text{New York},\,\textbf{2015-09-10})$} \\
\textbf{Sub-answer} & $t_1 = \textbf{2015-09-10}$ \\
\midrule
\textbf{Q2} & \textbf{When was the 2018 Summit held?} \\
\textbf{Selected Seed} & [2018 Summit] \\
\textbf{Indicator} & $(\text{2018 Summit},\,\text{held in},\,?y,\,t_2)$ \\
\textbf{Toolkit \& Params} & \texttt{OneHop}(entity = 2018 Summit, after = 2018-01-01, before = 2019-01-01) \\
\textbf{Retrieved Facts} & 
\textbf{$(\text{2018 Summit},\,\text{held\_in},\,\text{Singapore},\,\textbf{2018-11-22})$} \\
\textbf{Sub-answer} & $t_2 = \textbf{2018-11-22}$ \\
\midrule
\textbf{Q3} & \textbf{Between $t_1$ and $t_2$, which company collaborated with Microsoft?} \\
\textbf{Selected Seed} & [Microsoft] \\
\textbf{Indicator} & $(?z,\,\text{collaborate with},\,\text{Microsoft},\,t_3)$ \\
\textbf{Time vars} & $[\text{between}(t_3,[t_1,t_2])]$ \\
\textbf{Toolkit \& Params} & \texttt{BetweenRange}(entity = Microsoft, between = ($t_1$, $t_2$), relation\_filter = collaborate with) \\
\textbf{Retrieved Facts} &
\textbf{$(\textbf{NVIDIA},\,\text{collaborate\_with},\,\text{Microsoft},\,\textbf{2016-05-20})$;} \\ &
\textbf{$(\textbf{OpenAI},\,\text{collaborate\_with},\,\text{Microsoft},\,\textbf{2018-03-14})$;} \\ &
$(\text{Apple},\,\text{collaborate\_with},\,\text{Microsoft},\,\text{2019-02-10})$ \\
\textbf{Sub-answer} & \textbf{NVIDIA} (2016-05-20) and \textbf{OpenAI} (2018-03-14), both satisfying $t_1<t_3<t_2$ \\
\toprule
\textbf{Temporal Reasoning Chain} &
\textbf{2015 Conference} $\xrightarrow[\text{2015-09-10}]{\text{held\_in}}$ New York 
$\;\;\leadsto\;\;$ \textbf{NVIDIA} $\xrightarrow[\text{2016-05-20}]{\text{collaborate\_with}}$ \textbf{Microsoft} 
$\;\;\leadsto\;\;$ \textbf{OpenAI} $\xrightarrow[\text{2018-03-14}]{\text{collaborate\_with}}$ \textbf{Microsoft} 
$\;\;\leadsto\;\;$ \textbf{2018 Summit} $\xrightarrow[\text{2018-11-22}]{\text{held\_in}}$ Singapore. \\
\textbf{Temporal Faithfulness} & All facts satisfy $t_1 < t_3 < t_2$; valid interval reasoning confirmed. \\
\textbf{Final Answer} & \textbf{NVIDIA}, \textbf{OpenAI}. \\
\textbf{Response} &
“Anchoring the 2015 Conference at 2015-09-10 and the 2018 Summit at 2018-11-22, the companies collaborating with Microsoft within this interval are \textbf{NVIDIA} (2016) and \textbf{OpenAI} (2018). Both satisfy the temporal between constraint.” \\
\bottomrule
\end{tabularx}
\end{table*}

\clearpage
\newpage
\onecolumn

\section{Prompts}\label{appendix:prompt}
In this section, we detail the prompts required for our main experimental procedures.

\begin{center}
\begin{minipage}{0.9\columnwidth}
    \vspace{2mm}
    \centering
    \begin{tcolorbox}[title=Temporal Type Classification Prompt Template]
        \small
Given a natural-language question that may include explicit dates, relative temporal expressions, or comparative phrases, your task is to classify it into \text{exactly one} supported temporal type that best represents its temporal intent and logical operator.  
If multiple categories appear possible, choose the \text{most specific operator-sensitive} type.

Supported temporal types (single label only):  
\texttt{equal, before, after, during, between, first, last, beforeNlast, afterNfirst, count, comparison.}


\vspace{5pt}
\text{Experience Examples}:
\texttt{\{In-Context Few-shot\}}
\vspace{5pt}

Q: \{Question\}

A:
    \end{tcolorbox}
    \vspace{2mm}
\end{minipage}
\end{center}
\noindent
The \texttt{\{In-Context Few-shot\}} examples are retrieved from \text{experience memory}, representing previously successful classification traces.  
They guide the model in identifying the temporal operator patterns (e.g., “first Y after Z”) and selecting the most appropriate category.

\vspace{6mm}

\begin{center}
\begin{minipage}{0.9\columnwidth}
    \centering
    \vspace{2mm}
    \begin{tcolorbox}[title=Question Tree Construction Prompt Template]
        \small
Given a temporal question and its classified type \texttt{\{Question\_Type\}}, your task is to decompose it into a structured reasoning tree that contains complete subquestions, indicators, explicit temporal constraints, and time variables.  

Map the question to its most relevant decomposition pattern according to its temporal operator and structural characteristics.  
Generate subquestions that reflect key reasoning steps, construct indicators for entity–relation–time tuples,  
and specify explicit temporal constraints (e.g., \texttt{t2 > t1}, \texttt{before(t2, t1)}).  
Each subquestion must be a complete, grammatically correct sentence, and all time variables should preserve monotonic order (\(t_1 \leq t_2 \leq ... \leq t_n\)).  
Use \texttt{?x}, \texttt{?y} for unknown entities and \texttt{t1}, \texttt{t2} for time variables.  
Represent indicators as \texttt{Entity1 --[relation]--> Entity2 (t)}.  
Return the output strictly in the following format:

\begin{verbatim}
Subquestions: [sub1, sub2, ...]
Indicators:   [edge1, edge2, ...]
Constraints:  [constraint1, constraint2, ...]
Time_vars:    [t1, t2, ...]
\end{verbatim}


\vspace{5pt}
\text{Experience Examples}:
\texttt{\{In-Context Few-shot\}}
\vspace{5pt}

Q: \{Question\}

A:
    \end{tcolorbox}
\end{minipage}
\end{center}
\vspace{3mm}

\noindent
The \texttt{\{Question\_Type\}} is obtained from the temporal classification stage.  
Each retrieved example in \texttt{\{In-Context Few-shot\}} illustrates how similar questions were decomposed into substeps, indicators, and constraints, guiding the model to align with proven reasoning patterns.

\vspace{6mm}

\begin{center}
\begin{minipage}{0.9\columnwidth}
    \vspace{2mm}
    \centering
    \begin{tcolorbox}[title=Seed Selection Prompt Template]
        \small
Given a subquestion, its reasoning indicator, the topic entities, and optional contextual or temporal hints, your task is to select a concise set of \text{seed entities} to initialize retrieval.  
Prefer specific, high-yield entities that anchor reasoning paths; avoid overly broad categories.  
If multiple options exist, choose the \text{minimal} set that best covers the intended reasoning target.

\vspace{5pt}
\text{Experience Examples}:
\texttt{\{In-Context Few-shot\}}
\vspace{5pt}

Q: \{Subquestion\}

Think Indicator: \{Think\_Indicator\}

Available Topic Entities: \{Available\_Entities\}

Context Info (optional): \{Context\_Info\}

Time Hints (optional): \{Time\_Hints\}

A:
    \end{tcolorbox}
\end{minipage}
\end{center}
\vspace{3mm}

\noindent
Seed selection prompts guide the retrieval initialization phase in Section \ref{sec:retrieval}.  
The examples retrieved from memory illustrate successful entity anchors for similar subquestions, enabling the model to leverage prior reasoning experience when determining where to begin exploration.



\begin{center}
\begin{minipage}{0.9\columnwidth}
    \vspace{2mm}
    \centering
    \begin{tcolorbox}[title=Experience-Guided Toolkit Selection Prompt Template]
        \small
Given a subquestion, its reasoning indicator, temporal type, and contextual information, your task is to select the most suitable temporal toolkit(s) for solving it.  
Multiple toolkits may be selected if the reasoning requires combined temporal operations.

You should identify which toolkit(s) align with the temporal operator and reasoning goal of the question.  
Recommend the most appropriate toolkit configuration(s), specifying both parameters and reasoning rationale.

Available toolkits with descriptions: \{Available\_toolkits\}

Expected Output (JSON):
\begin{verbatim}
{ "selected_toolkits": [ {"original_name": "ToolkitName", "reasoning": "Reason statement", "priority": 1
"parameters": {"entity1": "...", "entity2": "...", "limit": "Number", ...}}] }
\end{verbatim}

\vspace{5pt}
\text{Experience Examples}:
\texttt{\{In-Context Few-shot\}}
\vspace{5pt}

Q: \{Subquestion\}

Think Indicator: \{Think\_Indicator\}

Temporal Type: \{Question\_Type\}

Seed Info: \{Seed\_Info\}

Time Hints: \{Time\_Hints\}

A:
    \end{tcolorbox}
\end{minipage}
\end{center}
\vspace{3mm}

\noindent
When a similar trace exists in the experience pool, MemoTime retrieves exemplar toolkit configurations as few-shot examples.  
Otherwise, the model enters a \text{cold-start} mode and infers a configuration from the question structure and temporal hints.  
Each selected toolkit represents a specific reasoning pattern (e.g., event ordering, interval comparison, timeline construction).

\vspace{6mm}

\begin{center}
\begin{minipage}{0.9\columnwidth}
    \vspace{2mm}
    \centering
    \begin{tcolorbox}[title=Multiple Toolkits Debate–Vote Prompt Template]
        \small

Given a subquestion and the outputs generated by multiple temporal toolkits, your task is to evaluate and compare their results to determine which toolkit provides the most \text{accurate}, \text{temporally faithful}, and \text{semantically consistent} answer. 
Each toolkit represents a distinct reasoning or retrieval strategy, and their outputs may vary in completeness, reliability, and explanatory depth.

The evaluation process compares all toolkit outputs across several complementary dimensions.  
Each candidate answer is assessed for its \text{relevance} to the subquestion, \text{temporal faithfulness} to the given time constraints, and \text{path validity} within the temporal knowledge graph.  
The model further considers \text{evidence completeness}, \text{semantic consistency}, and the \text{adequacy} of each toolkit’s configuration and parameters.  
When multiple valid but conflicting answers arise, preference is given to the one exhibiting stronger temporal grounding, clearer provenance, and greater explanatory transparency.  
Through this comparative reasoning, the model identifies the toolkit whose output demonstrates the most coherent, temporally consistent, and well-supported reasoning chain.

\text{Expected Output (JSON):}
\begin{verbatim}
{
    "winning_toolkit": <integer>,
    "winning_answer": {
        "entity": "<string>", "time": "<YYYY[-MM[-DD]] or Unknown>",
        "path": ["head", "relation", "tail"], "score": <float>, "reason": "<concise selection rationale>"
  },
  "evaluation": { 
    "criteria_scores": {
        "toolkit_1": {"relevance": <0-1>, "accuracy": <0-1>, "completeness": <0-1>} 
        "toolkit_2": {"relevance": <0-1>, "accuracy": <0-1>, "completeness": <0-1>} 
        ...//include all evaluated toolkits },
    "overall_winner": "<Toolkit number>", "reasoning": "<short comparative analysis across toolkits>" }
}
\end{verbatim}

\vspace{5pt}
\text{Successful Examples:}  
\texttt{\{In-Context Few-shot\}}
\vspace{5pt}

Q: \{Subquestion\}  

Toolkit Results: \{Collected\_Results\}  

A:
    \end{tcolorbox}
    \vspace{2mm}
\end{minipage}
\end{center}

\noindent
This prompt is used at the \text{debate–vote stage}, where MemoTime aggregates and evaluates the outputs from all executed temporal toolkits.  
The model receives structured evidence from each toolkit—including reasoning paths, timestamps, parameters, and explanatory notes—and performs comparative reasoning according to the defined evaluation process.  
The decision prioritizes \text{temporal faithfulness}, \text{evidence completeness}, and \text{semantic consistency}, ensuring that the final selected answer is both contextually grounded and logically coherent across all candidate toolkits.
\vspace{3mm}

\begin{center}
\begin{minipage}{0.9\columnwidth}
    \centering
    \begin{tcolorbox}[title=LLM-aware Selection Prompt Template]
        \small
Given the main question, a chain of thought generated by the LLM that considers all entities, a set of split subquestions, and their retrieved knowledge graph paths, your task is to score the candidate paths and identify the \text{top three} that are most likely to contain the correct evidence for the question.

\vspace{5pt}
\text{Successful Examples}:
\texttt{\{In-Context Few-shot\}}
\vspace{5pt}

Q: \{Query\}

Think Indicator: \{Think\_Indicator\}

Candidate Paths: \{Candidate\_Paths\}

A:
    \end{tcolorbox}
\end{minipage}
\end{center}
\vspace{3mm}

\noindent
This prompt lets the LLM act as a reasoning critic, ranking candidate paths using patterns learned from previous reasoning examples.  
The \texttt{\{In-Context Few-shot\}} samples come from successful path selections in prior tasks, demonstrating how to evaluate relevance, temporal ordering, and completeness.

\begin{center}
\begin{minipage}{0.9\columnwidth}
\vspace{2mm}
\centering
\begin{tcolorbox}[title=Sufficiency Evaluation Prompt Template]
\small
Given a question (or subquestion), its candidate answer(s), retrieved evidence paths, and reasoning context, your task is to determine whether the information provided is \text{sufficient and consistent} to answer the target question.

You should evaluate in two modes:
\begin{itemize}[leftmargin=10pt]
    \item \textbf{Local (Subquestion) Sufficiency}: Assess whether the retrieved evidence and reasoning path for this subquestion are adequate to support the proposed answer.  
    If sufficient, respond \{True\} and restate the answer; otherwise respond \{False\}, provide a short explanation, and specify one corrective action from \texttt{\{Decompose, Refine, Retrieval Again\}}.
    \item \textbf{Global (Full Question) Sufficiency}: Assess whether the entire reasoning trajectory, including all subquestions, intermediate answers, and aggregated evidence, is coherent and complete enough to answer the main question.  
    If sufficient, respond \{True\} and provide the final answer; otherwise respond \{False\} with a reason and one corrective action from the same set.
\end{itemize}

\vspace{5pt}
\text{Successful Examples:}  
\texttt{\{In-Context Few-shot\}}

\vspace{5pt}
Q: \{Question / Subquestion\}  

Candidate Answer(s): \{Answer\_Entity / Final\_Answer\}  

Evidence Paths: \{Evidence\_Paths\}  

Reasoning Context / Trajectory: \{Subquestions\_and\_Answers\}  

A:
\end{tcolorbox}
\end{minipage}
\end{center}

\vspace{3mm}

\noindent
This unified prompt template supports both \text{local} and \text{global} sufficiency evaluation.  
At the local level, it determines whether each reasoning step is self-contained and evidence-supported.  
At the global level, it verifies that the overall reasoning chain forms a temporally consistent, logically complete path leading to the correct final answer.
\begin{center}
\begin{minipage}{0.9\columnwidth}
    \vspace{2mm}
    \centering
    \begin{tcolorbox}[title=Question Answering Generation Prompt Template]
        \small
Given the main question, a reasoning chain of thought think indicator, and the complete reasoning global trajectory (all solved split questions, step answers, aggregated reasoning paths, and used toolkits for each split question),  
your task is to generate the final answer using the provided knowledge paths and your own reasoning.
You should ensure that the final answer is logically consistent with the reasoning trajectory and fully supported by the retrieved paths.  
If multiple candidate entities are equally valid (e.g., simultaneous events or co-occurring facts), list all of them.

\vspace{5pt}
\text{Successful Examples}:
\texttt{\{In-Context Few-shot\}}
\vspace{5pt}

Q: \{Main\_Question\}

Think Indicator: \{Think\_Indicator\}

Global Trajectory Summary: \{SolvedQuestions,\_Answers,\_and\_Proofs\}

A:
    \end{tcolorbox}
\end{minipage}
\end{center}

\vspace{3mm}
\noindent
This final prompt synthesizes all previous reasoning and evidence into a natural-language answer.  
The exemplars demonstrate how previous successful reasoning trajectories were concluded, ensuring the generated answer remains temporally faithful, complete, and well-grounded.

\end{document}